\documentclass{article}

\usepackage{microtype}
\usepackage{graphicx}
\usepackage{subcaption}
\usepackage{booktabs} % for professional tables

\usepackage{hyperref}

\usepackage[preprint]{icml2026}

\usepackage{amsmath}
\usepackage{amssymb}
\usepackage{mathtools}
\usepackage{amsthm}

\usepackage{multirow}
\usepackage{colortbl}
\usepackage{adjustbox}
\usepackage{booktabs}
\usepackage{wrapfig}
\usepackage{graphicx}
\usepackage{amsmath}
\usepackage{amssymb}
\usepackage{mathtools}
\usepackage{amsthm}
\usepackage{multirow}
\usepackage{colortbl}
\usepackage{adjustbox}
\usepackage{tcolorbox}

\definecolor{maroonx}{RGB}{195,18,48}
\definecolor{darkred}{RGB}{192, 0, 0}
\definecolor{darkgreen}{RGB}{103, 174, 64}
\definecolor{darkblue}{RGB}{56, 84, 146}
\definecolor{graybg}{gray}{0.90}

\usepackage[capitalize,noabbrev]{cleveref}

\theoremstyle{plain}

\theoremstyle{definition}

\theoremstyle{remark}

\usepackage[textsize=tiny]{todonotes}

\icmltitlerunning{Towards Interpretable Hallucination Analysis and Mitigation in LVLMs via Contrastive Neuron Steering}

\begin{document}

\twocolumn[
  \icmltitle{
  Towards Interpretable Hallucination Analysis and Mitigation in LVLMs via Contrastive Neuron Steering
  }

\begin{icmlauthorlist}
  \icmlauthor{Guangtao Lyu}{xdu1}
  \icmlauthor{Xinyi Cheng}{xdu2}
  \icmlauthor{Qi Liu}{xdu1}
  \icmlauthor{Chenghao Xu}{hhu}
  \icmlauthor{Jiexi Yan}{xdu2}
  \icmlauthor{Muli Yang}{i2r}
  \icmlauthor{Fen Fang}{i2r}
  \icmlauthor{Cheng Deng}{xdu1}
\end{icmlauthorlist}

% ===== 单位定义 =====
\icmlaffiliation{xdu1}{School of Electronic Engineering, Xidian University, Xi'an, China}
\icmlaffiliation{xdu2}{School of Computer Science and Technology, Xidian University, Xi'an, China}
\icmlaffiliation{hhu}{College of Computer and Information, Hohai University, Nanjing, China}
\icmlaffiliation{i2r}{Institute for Infocomm Research, A*STAR, Singapore}

% ===== 通讯作者（Cheng Deng）=====
\icmlcorrespondingauthor{Cheng Deng}{chdeng.xd@gmail.com}

  \vskip 0.3in
]

% \author{Guangtao Lyu$^{1}$, Xinyi Cheng$^{2}$, Chenghao Xu$^{3}$, Qi Liu$^{1}$, Muli Yang$^{4}$, Fen Fang$^{4}$, \\ Huilin Chen$^{5}$, Jiexi Yan$^{2}$,  Xu Yang$^{1}$, Cheng Deng$^{1}$\thanks{Corresponding author} \\
%         $^{1}$ School of Electronic Engineering, Xidian University, China, 
%         $^{2}$ School of Computer Science and Technology, \\  Xidian University, China,  $^{3}$ Hohai university, China, 
%         $^{4}$ Institute for Infocomm Research (I\textsuperscript{2}R), A*STAR, \\ Singapore,  
%         $^{5}$School of Foreign Languages, Xidian University, China, \\
%         \texttt{ \{guangtaolyu,qiliu,xinyicheng\}@stu.xidian.edu.cn,  fang fen@a-star.edu.sg} ,\\ 
%         \texttt{\{jxyan1995,muliyang.xd,xuyang.xd,chdeng.xd\}@gmail.com, hlchen@xidian.edu.cn} }



\printAffiliationsAndNotice{}  % no special notice (required even if empty)

\begin{abstract}
LVLMs achieve remarkable multimodal understanding and generation but remain susceptible to hallucinations. Existing mitigation methods predominantly focus on output-level adjustments, leaving the internal mechanisms that give rise to these hallucinations largely unexplored.
To gain a deeper understanding, we adopt a representation-level perspective by introducing sparse autoencoders (SAEs) to decompose dense visual embeddings into sparse, interpretable neurons. Through neuron-level analysis, we identify distinct neuron types, including always-on neurons and image-specific neurons. Our findings reveal that hallucinations often result from disruptions or spurious activations of image-specific neurons, while always-on neurons remain largely stable. Moreover, selectively enhancing or suppressing image-specific neurons enables controllable intervention in LVLM outputs, improving visual grounding and reducing hallucinations.
Building on these insights, we propose Contrastive Neuron Steering (CNS), which identifies image-specific neurons via contrastive analysis between clean and noisy inputs. CNS selectively amplifies informative neurons while suppressing perturbation-induced activations, producing more robust and semantically grounded visual representations. This not only enhances visual understanding but also effectively mitigates hallucinations. By operating at the prefilling stage, CNS is fully compatible with existing decoding-stage methods.
Extensive experiments on both hallucination-focused and general multimodal benchmarks demonstrate that CNS consistently reduces hallucinations while preserving overall multimodal understanding.
\end{abstract}

\section{Introduction}\label{sec:intro}

Large vision-language models (LVLMs)~\cite{llava,instructblip,qwenvl,zhu2023minigpt4} have achieved remarkable progress in multimodal understanding and generation. Despite these advances, LVLMs remain vulnerable to \emph{hallucinations}, particularly object hallucinations where the model describes entities that are not present in the input image~\cite{lee2018hallucinations,vcd}. Such errors undermine reliability and user trust, while raising critical concerns for safety-sensitive applications such as autonomous systems, medical imaging, and decision support.

To mitigate hallucinations, numerous techniques have been investigated, including visual instruction fine-tuning~\cite{liu2024llava1.5,yu2024hallucidoctor}, integration with external expert models, and contrastive decoding strategies~\cite{vcd,halc,m3id}. Nevertheless, the mechanistic origins of hallucinations remain poorly understood. Existing explanations predominantly attribute hallucinations to language biases, such as the “anchor pattern”~\cite{opera} and “text inertia”~\cite{pai}, which posit that hallucinations emerge from the dominance of linguistic priors over visual features. However, these perspectives largely neglect the internal visual representation space of LVLMs. In this paper, we seek to explore the relationship between internal visual representations and hallucinations, addressing the following fundamental questions: 
how are visual features organized internally, how they change under perturbations, and which aspects of the representation most directly contribute to hallucinations?

\begin{figure}[t]
    \centering
    \includegraphics[width=1\linewidth]{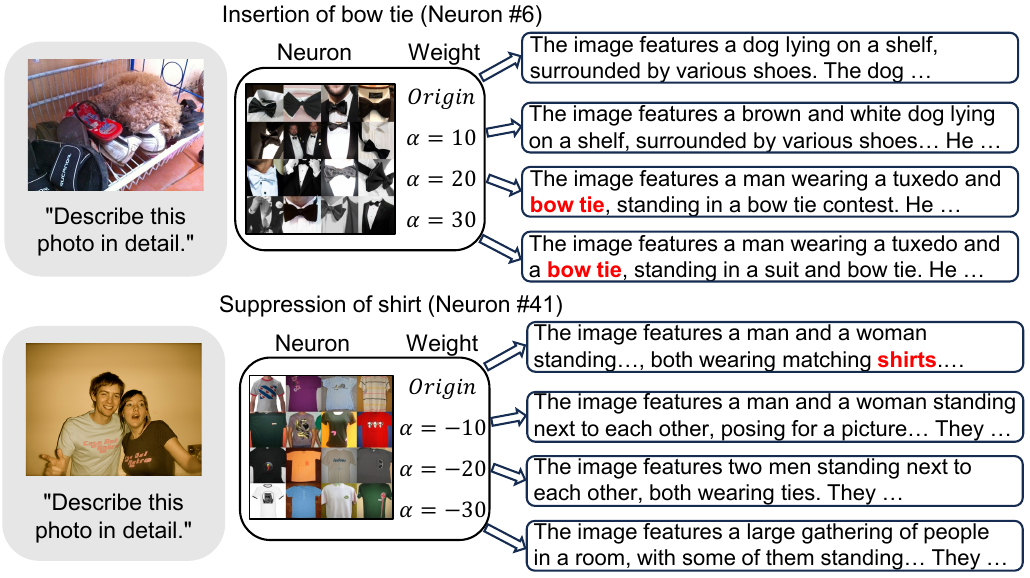}
    \vspace{-8pt}
\caption{Steering an LVLM: (a) amplifying a ``bow tie" neuron emphasizes this concept in generated descriptions, while (b) suppressing a ``shirt" neuron prevents it from appearing.}
    \vspace{-8pt}

    \label{fig:single_neuron_steering}
\end{figure}

To investigate hallucinations in depth, we adopt a representation-level perspective by decomposing dense, entangled visual features from LVLM visual encoders into sparse, interpretable neurons using sparse autoencoders (SAEs)~\cite{makhzani2013k,templeton2024scaling}. These neurons capture concept-specific visual features~\cite{durmus2024steering,templeton2024scaling}, providing a principled basis for analyzing the internal mechanisms of hallucinations (\Cref{fig:single_neuron_steering}). Neuron-level characterization reveals two distinct types: a small set of \emph{always-on neurons}, which remain active across all images and encode semantically irrelevant information, and a larger set of \emph{image-specific neurons}, whose sparse, semantically grounded activations correspond directly to specific visual concepts (\Cref{fig:image_level_neurons_vis,fig:patch_level_neuron_vis}). This distinction allows us to systematically investigate how disruptions or spurious activations of image-specific neurons contribute to hallucinations and to explore targeted interventions that modulate LVLM outputs in a controlled and interpretable manner.

Building on the sparse neuron representation, we systematically analyze hallucinations in LVLMs from an interpretable, internal perspective. We find that hallucinations primarily arise from spurious activations or disruptions of \emph{image-specific neurons}, which encode semantically meaningful visual concepts, while \emph{always-on neurons} remain stable and have limited influence on task-specific predictions. By tracing neuron activations across images and tasks, we identify the internal failure modes that lead to hallucinated outputs.  We further investigate neuron-level interventions and show that selectively amplifying relevant image-specific neurons or suppressing misleading ones enables controlled modulation of model outputs(~\Cref{fig:case_sheep,fig:case_black_apple,fig:steering_two_neurons}). Overall, this neuron-level analysis provides both an interpretable explanation of hallucinations and a principled mechanism for intervention, improving visual understanding and effectively mitigating hallucinations in LVLMs.

Building on these insights, we propose a novel method, {Contrastive Neuron Steering (CNS)}, to mitigate hallucinations by operating directly in the internal visual representation space. CNS identifies image-specific neurons through contrastive analysis between clean and noisy inputs and selectively amplifies informative neurons while suppressing perturbation-induced activations. To further reduce interference from neurons that are persistently active across images, CNS incorporates an \emph{Always-on Neuron Suppression (ANS)} mechanism, which down-weights these non-informative signals and sharpens focus on image-specific features. By producing more robust and semantically grounded visual representations, CNS enhances visual understanding and effectively mitigates hallucinations. Operating at the prefilling stage, CNS is fully compatible with existing decoding-stage methods, providing a general and complementary solution to prior output-level approaches.

In summary, our contributions are as follows:
\begin{itemize}
\item We leverage SAEs to decompose dense visual representations into sparse, interpretable neurons, providing an internal, mechanistic perspective for analyzing and intervening in LVLMs hallucinations.
\item We show that hallucinations are primarily caused by disruptions or spurious activations of image-specific neurons, with always-on neurons having limited influence, and that targeted interventions on image-specific neurons can effectively modulate model outputs and mitigate hallucinations.

\item We propose Contrastive Neuron Steering (CNS), a representation-level method that mitigates hallucinations by selectively enhancing image-specific neurons via contrastive analysis of clean and noisy inputs. CNS operates at the prefilling stage and is compatible with existing decoding-stage techniques.

\item Extensive experiments on both hallucination and general multimodal benchmarks show that CNS effectively improves visual grounding and reduces hallucinations, while preserving overall multimodal understanding.
\end{itemize}

\section{Related Work}
\noindent \textbf{Hallucinations in LVLMs.}
LVLMs~\cite{openai2024gpt4technicalreport,anthropic2024claude,deepseekai2025deepseekr1incentivizingreasoningcapability,comanici2025gemini,yang2025qwen3} have achieved significant progress in multimodal understanding and generation. However, these models remain prone to hallucinations, particularly object hallucinations~\cite{liu2024survey,lee2023volcano,gunjal2024detecting,dola}. The causes include pretraining data biases~\cite{agarwal2020towards,agrawal2016analyzing}, over-reliance on parametric knowledge~\cite{vcd,lee2023volcano,zhibo2023overcoming}, and biased visual feature learning~\cite{zhu2024ibd,opera,yue2024less,han2022visual}.
Existing hallucination mitigation strategies fall into two categories: training-driven methods, which fine-tune LVLMs via data augmentation or reinforcement learning~\cite{liu2023mitigating,sun2023aligning,zhou2024aligning,liu2024llava1.5,zhai2024halleControl}, and training-free methods, which intervene during inference by manipulating attention maps or logits using contrastive decoding or auxiliary models~\cite{hallu_attention_lens,Seeing_but_Not_Believing,dola_looking_twice_memvr_ffn_iternal,lyu2025revealing_guangtao_lmm,second,icd,cd,kaduri2025s_cvpr25}.

\noindent \textbf{Sparse Autoencoders.}  
SAEs~\cite{templeton2024scaling,monosamantic,sae_survey} decompose hidden activations into sparse neurons, providing an interpretable basis for analyzing and steering LVLMs. Recent advances improve sparsity and reconstruction, including BatchTopK~\cite{bussmann2024batchtopk}, JumpReLU~\cite{rajamanoharan2024jumping}, and Matryoshka variants~\cite{nabeshima2024matryoshka,bussmann2024matryoshka}. 
In LLMs, SAEs have been applied to neuron-level steering for reducing toxicity, sycophancy, and refusal~\cite{gallifant2025sparse,nanda2024progress}, as well as hallucination detection and safety~\cite{ferrando2025do,demircan2025sparse,wu2025interpreting}.  
In vision and multimodal settings, SAEs enable fine-grained interpretation and control: Revelio~\cite{Revelio} uncovers features in diffusion models, Matryoshka SAEs balance sparsity and reconstruction on CLIP embeddings~\cite{bussmann2024matryoshka}, USAEs align concepts across networks~\cite{thasarathan2025universal}, and SAE-V~\cite{lou2025sae} supports cross-modal interpretation and steering~\cite{zhang2024large}.

\begin{figure}[t]
    \centering
    \includegraphics[width=1\linewidth]{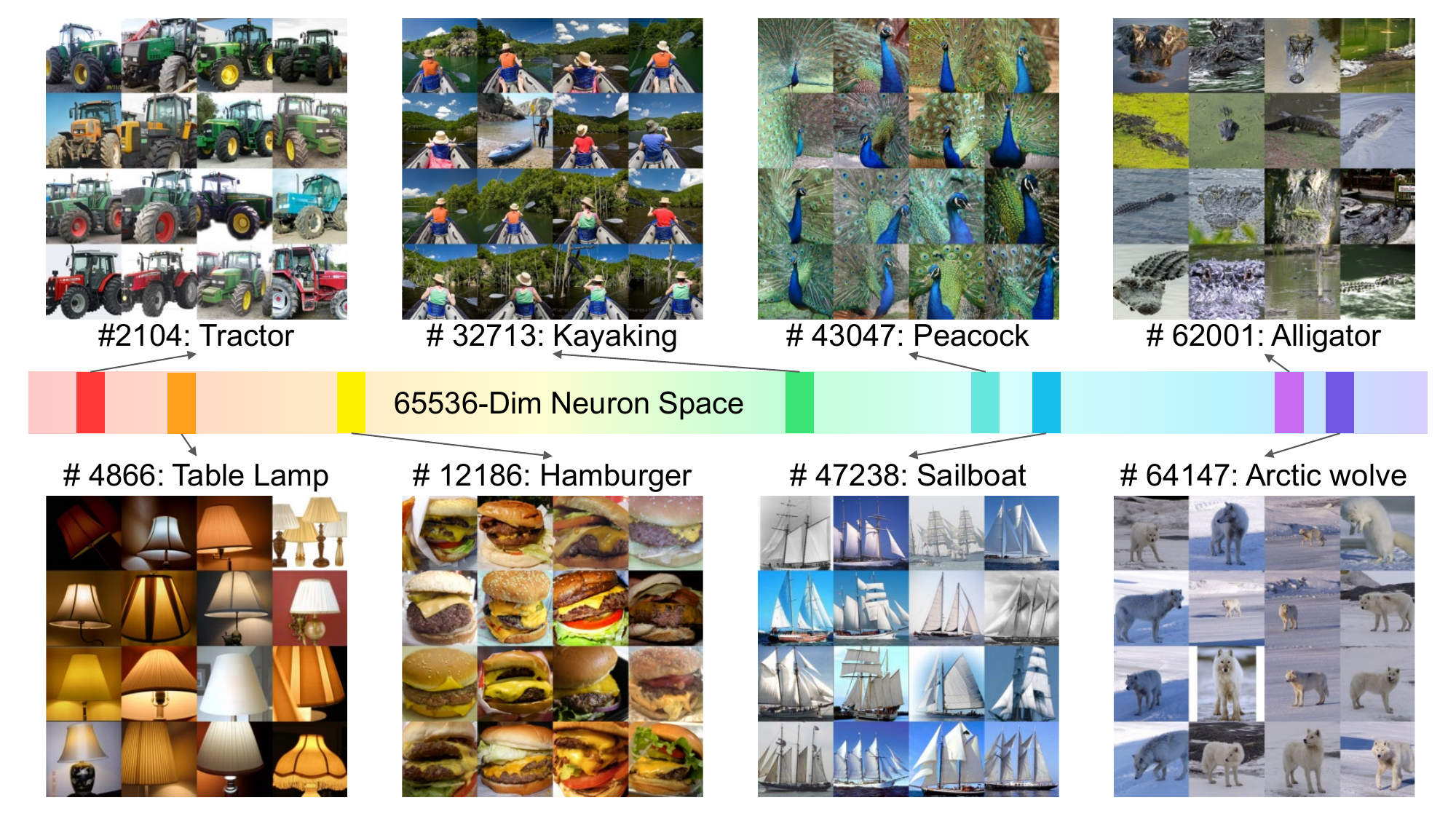}
    \vspace{-12pt}
\caption{Neuron visualizations from SAE, showing diverse visual patterns and semantic structures.}
    %%\vspace{-12pt}
    \label{fig:neuron_vis}
\end{figure}

\begin{table}[t]
    \vspace{-4pt}
\caption{Effects of zeroing different types of SAE neurons.}
    \vspace{-4pt}
\label{tab:zeros_neuron_types}
\centering
\resizebox{0.95\linewidth}{!}{
\begin{tabular}{lcc}
\toprule
\textbf{Neuron Type} & \textbf{Accuracy (\%)} $\uparrow$ & \textbf{F1-score (\%)} $\uparrow$ \\
\midrule
baseline        & 84.63 & 84.99 \\
\hline
always-on       & {84.68} & 85.08 \\
image-specific  & 63.08 & 57.36 \\
random          & 84.31 & 84.65 \\
\bottomrule
\end{tabular}}
\end{table}

\begin{figure}[t]
    \centering
    \includegraphics[width=1\linewidth]{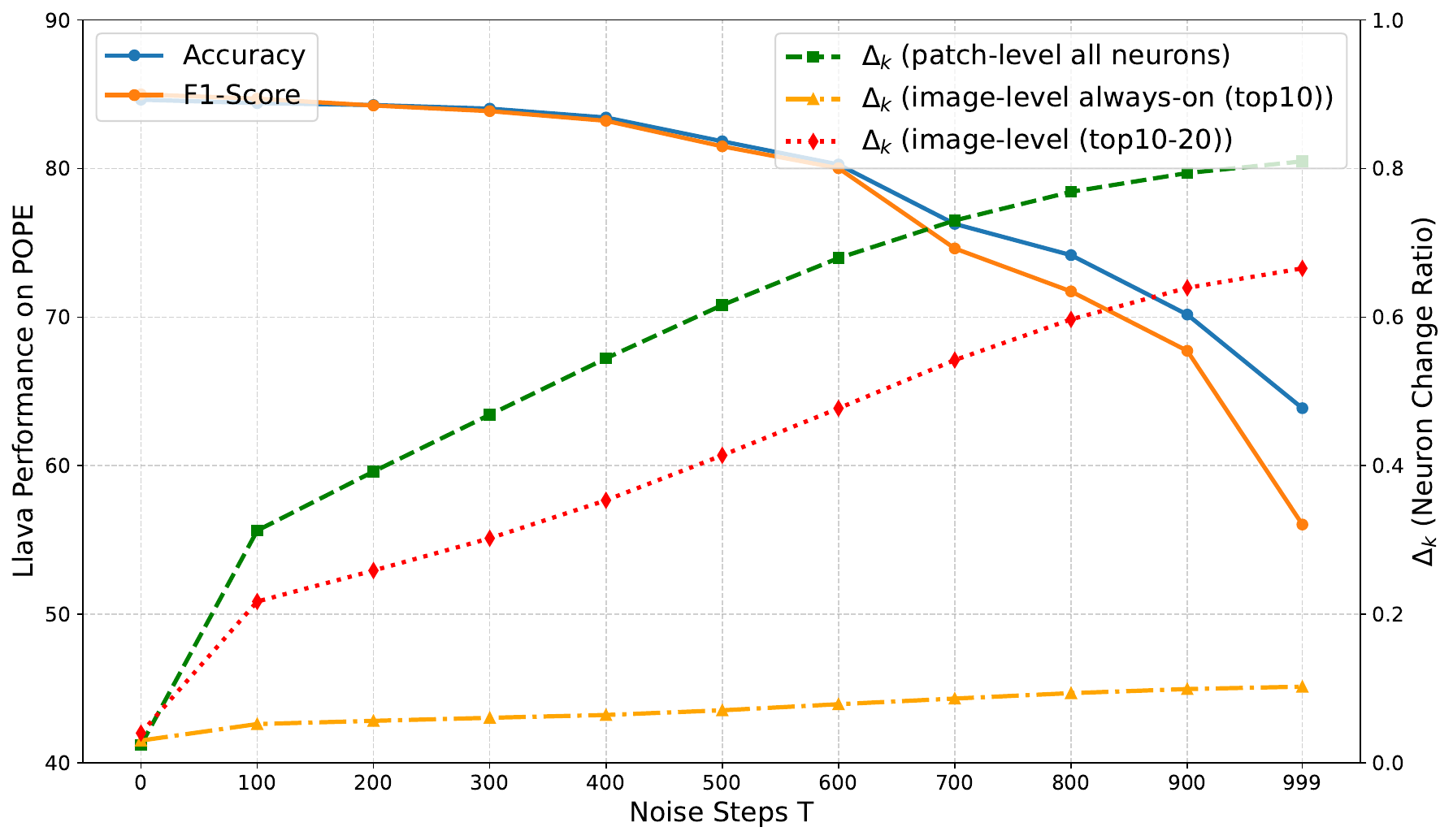}
    \vspace{-16pt}
\caption{Relationship between noise step, model performance, and activation changes across different neuron types.}
    \vspace{-8pt}
    \label{fig:explore_neuron_change}
\end{figure}

\section{Preliminaries}

\noindent \textbf{Sparse Autoencoders.}
SAEs are designed to disentangle dense and highly entangled internal representations by decomposing them into a sparse set of interpretable latent neurons~\cite{olshausen1997sparse, bricken2023towards}. 
Concretely, given a dense visual feature $\mathbf{v} \in \mathbb{R}^d$, the SAE encoder projects it into a latent space and enforces sparsity through a Top-$K$ constraint:
\begin{equation}
z(\mathbf{v}) = \mathrm{TopK}\big(\mathrm{ReLU}(W_{\mathrm{enc}}\mathbf{v} - \mathbf{b})\big),
\end{equation}
where only the $K$ largest activations are retained. 
The decoder then reconstructs the original feature as a linear combination of the activated latent neurons:
\begin{equation}
\hat{\mathbf{v}} = W_{\mathrm{dec}}^\top z(\mathbf{v}) + \mathbf{b}.
\end{equation}

\noindent \textbf{Inserting SAEs into LVLM.}
SAEs can be applied at various LVLM stages, including intermediate LLM layers or the visual encoder. Here, we focus on the \emph{visual encoder} for the following reasons:
(1) It enables fine-grained analysis of internal visual representations by isolating the visual encoder’s contribution, unlike prior studies that focus on strong language priors, allowing direct investigation of how visual feature changes relate to hallucinations.
(2) Some LVLMs adopt the same visual backbones, so an SAE trained on one backbone can be reused across LVLMs that share it.  
(3) Our approach operates at the \emph{prefilling stage} by regulating visual representations before decoding, while most existing methods act at the decoding stage, making it naturally compatible with and complementary to decoding-level approaches.
(4) Inserting SAEs into the visual encoder requires only one extra visual encoder forward, avoiding full model re-inference.

\begin{figure}[t]
    \centering
    \includegraphics[width=0.92\linewidth]{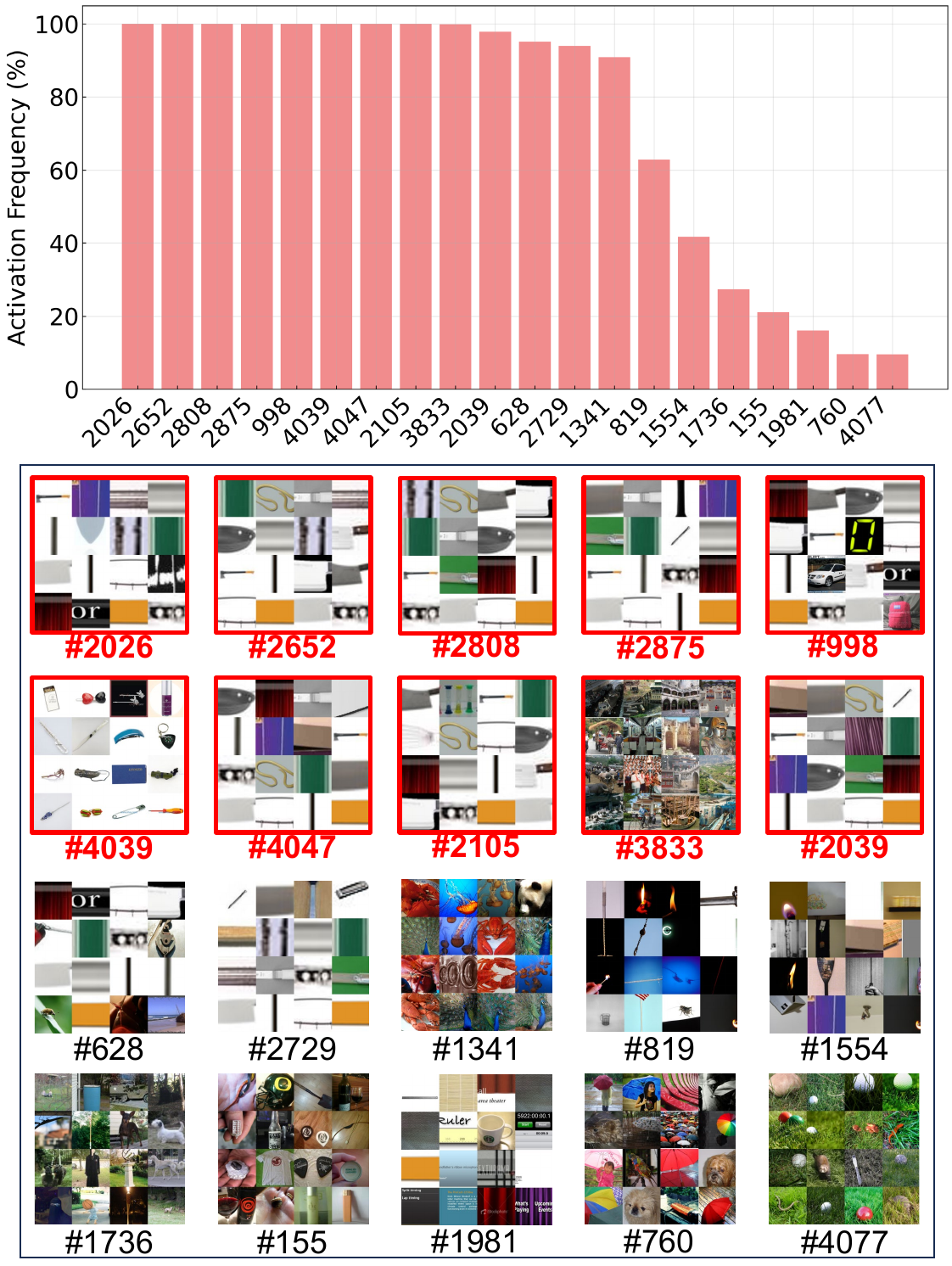}
    \vspace{-4pt}
    \caption{Image-level neuron analysis. Top: statistical analysis of neuron activations. Bottom: neuron visualization.  \textcolor{red}{Red boxes} indicate \emph{always-on neurons}, in the Top-20 for all images.}
    \vspace{-12pt}
    \label{fig:image_level_neurons_vis}
\end{figure}

\begin{figure}[t]
    \centering
    \includegraphics[width=0.92\linewidth]{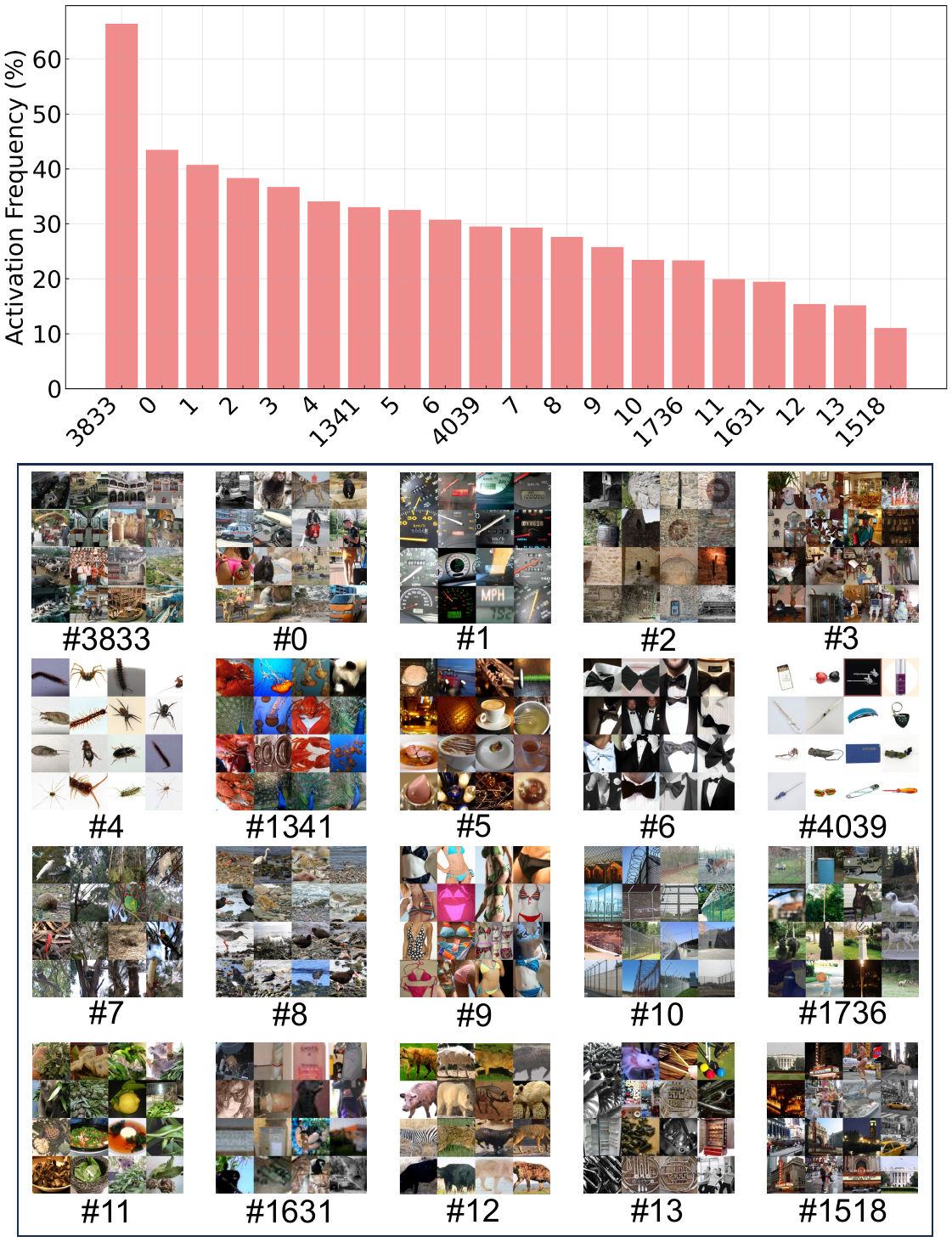}
    \vspace{-4pt}
    \caption{Patch-level neuron analysis. Top: statistical analysis of neuron activations. Bottom: neuron visualization.}
    \vspace{-8pt}
    \label{fig:patch_level_neuron_vis}
\end{figure}

\section{Interpretable Neuron-Level Hallucination Analysis and Mitigation}
\label{sec:analyse_sae}

We leverage neuron-level interpretability to systematically investigate the mechanisms underlying hallucinations in LVLMs and to explore strategies for mitigation.
We train a Matryoshka SAE~\cite{bussmann2024matryoshka} on image features extracted from the visual encoder of LLaVA-1.5 using ImageNet, and perform all analyses on the COCO dataset.

\subsection{Neuron Characterization}

To investigate the semantic structure captured by SAEs, we first visualize individual neurons by selecting, for each neuron, the top-16 images that trigger the strongest activations (\Cref{fig:neuron_vis,fig:more_neurons_vis}). This confirms that SAEs decompose dense visual representations into sparse, interpretable units, providing a foundation for subsequent mechanistic analysis and neuron-level interventions.

To systematically characterize neuron behavior, we combine quantitative and qualitative analyses at both the image and patch levels (\Cref{fig:patch_level_neuron_vis,fig:image_level_neurons_vis}). Quantitatively, we record the Top-20 most activated neurons for each image or spatial patch to capture activation statistics. Qualitatively, we visualize these Top-20 neurons to examine the semantic concepts they encode. This dual approach enables a thorough assessment of neuron behavior across spatial scales.

Through this analysis, we identify two distinct neuron types with characteristic properties. A small set of \emph{Always-on neurons} (10 out of 65k) consistently appear in the Top-20 across all images (red box in \Cref{fig:image_level_neurons_vis}). The images activating these neurons are visually similar and primarily reflect general textures or color regions, without encoding specific semantic content. In contrast, \emph{image-specific neurons} exhibit high selectivity: they are strongly activated by visually similar images (e.g., cats or grass) and capture semantically meaningful, localized concepts at the patch level (\Cref{fig:patch_level_neuron_vis}). This categorization provides the basis for all subsequent neuron-level analyses and interventions.

\begin{figure}[t]
    \centering
    \includegraphics[width=0.95\linewidth]{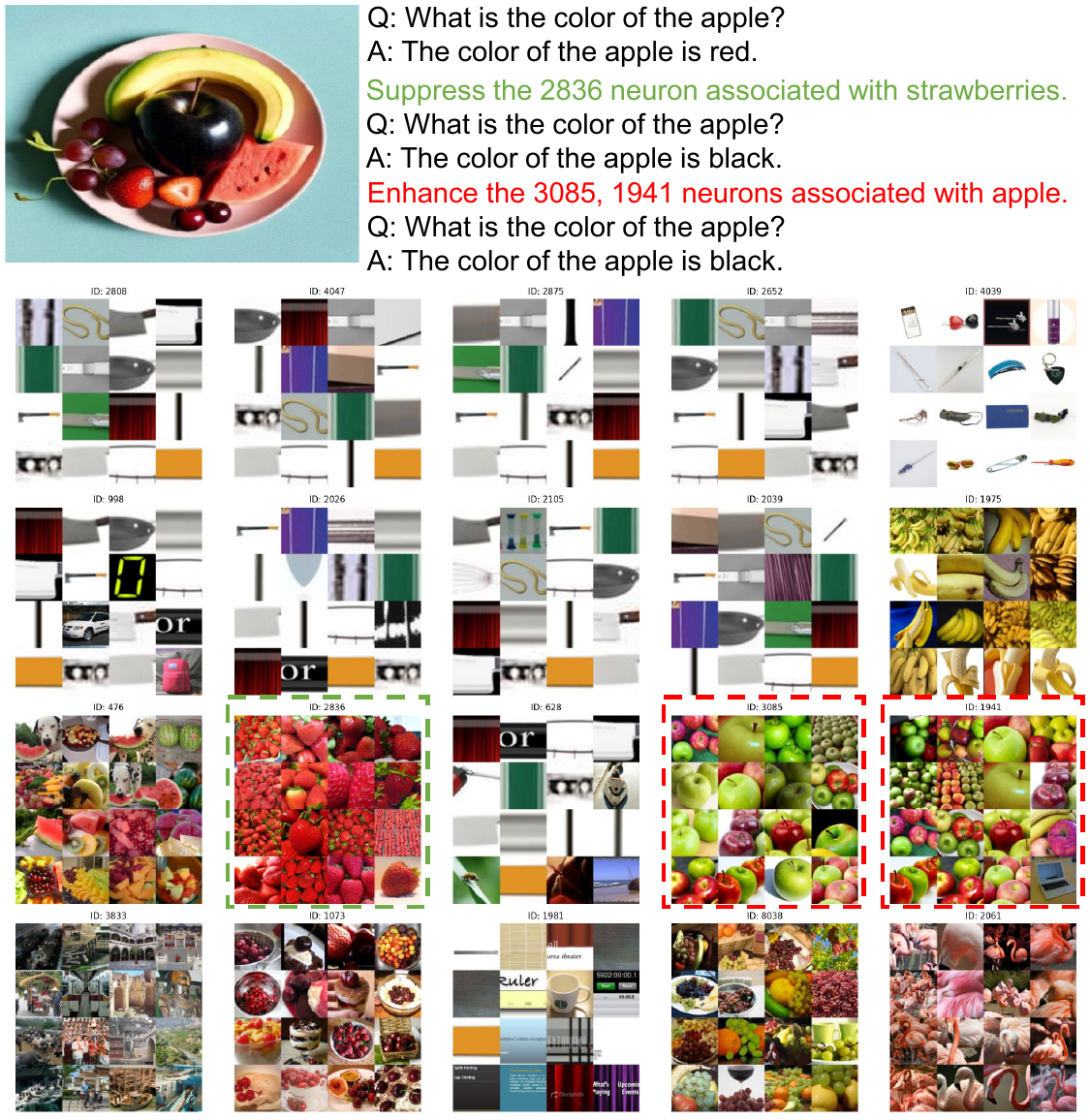}
    \vspace{-4pt}
\caption{Hallucination caused by suppression of relevant neurons. Strong activation of a strawberry-related neuron (\textcolor{green}{green boxes}) overwhelms apple-specific neurons (\textcolor{red}{red boxes}), leading to an incorrect color prediction. Targeted neuron steering restores correct visual grounding.}
    \vspace{-12pt}
    \label{fig:case_black_apple}
\end{figure}

\begin{figure}[t]
    \centering
    \includegraphics[width=0.95\linewidth]{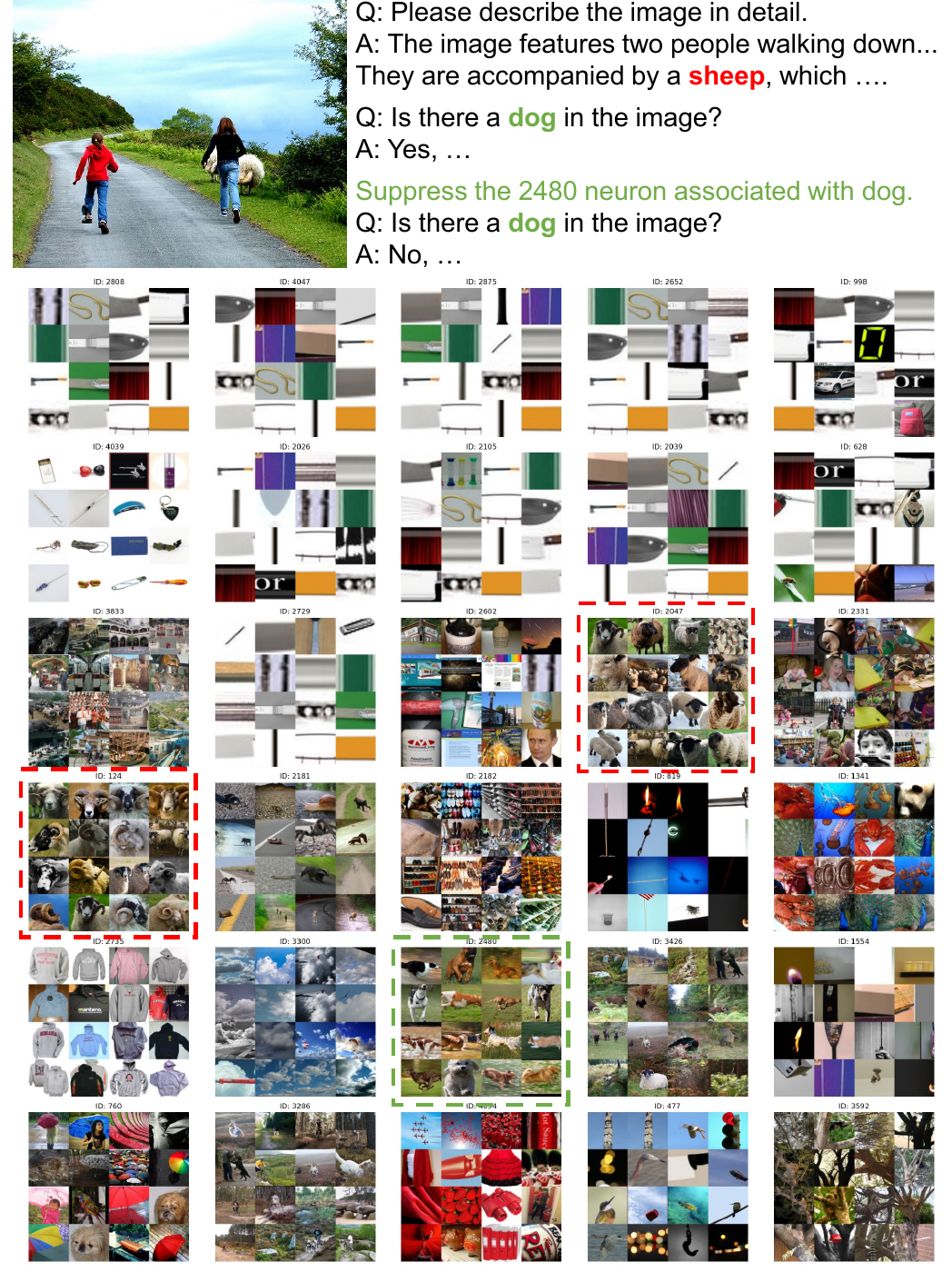}
    \vspace{-4pt}
\caption{Hallucination caused by spurious activation of irrelevant concepts. Sheep-related neurons dominate (\textcolor{red}{red boxes}), but additional activation of a dog-related neuron (\textcolor{green}{green boxes}) leads to a hallucinated response to a dog-specific query. Suppressing this neuron corrects the prediction.}
    \vspace{-8pt}
    \label{fig:case_sheep}
\end{figure}

\subsection{Neuron-Guided Analysis of Hallucinations}

Previous studies~\cite{vcd,only} have shown that adding image noise increases visual uncertainty and amplifies hallucinations. To investigate the underlying mechanisms, we analyze how different neuron types respond to such perturbations, focusing on always-on and image-specific neurons. Using the interpretable latent space provided by SAEs, we examine how noise alters internal visual representations. Experiments are conducted on LLaVA-1.5 using the POPE benchmark (COCO, random setup), tracking both performance degradation and neuron activation changes under increasing noise levels.

To quantify neuron-level instability, we measure the \emph{Top-$K$ neuron change ratio} between a clean image $v$ and its noisy counterpart $\tilde{v}$:
\begin{equation}
\Delta_K(v, \tilde{v}) = 1 - \frac{|z(v) \cap z(\tilde{v})|}{K},
\end{equation}
where $z(\cdot)$ denotes the set of Top-$K$ activated neurons. Higher $\Delta_K$  indicate greater disruption of visual features.

As shown in \Cref{fig:explore_neuron_change}, increasing noise intensity leads to both higher hallucination rates and larger neuron-change ratios. Patch-level neurons change most rapidly, reflecting their sensitivity to local perturbations. Always-on neurons remain largely stable, whereas image-specific neurons exhibit substantial shifts, approaching the magnitude observed at the patch level. These results suggest that hallucinations are primarily driven by disruptions to image-specific neurons rather than global visual features. Qualitative examples further illustrate this effect. For instance, when noise is added to an image of a camera, activations of camera-related image-specific neurons gradually decrease, causing the model to produce progressively less detailed descriptions until camera-related content is omitted entirely (\Cref{fig:add_noise_vis}).  

We further investigate the impact of interventions on different neuron types by selectively zeroing their activations to assess their influences in hallucinations. We compare three groups: always-on neurons, image-specific neurons, and randomly selected neurons beyond the Top-20. As shown in \Cref{tab:zeros_neuron_types}, suppressing image-specific neurons substantially alters model predictions and significantly increases hallucinations, whereas zeroing always-on or random neurons has minimal effect. These results confirm that image-specific neurons are the primary drivers of hallucinations. 

\begin{figure}[t]
    \centering
    \includegraphics[width=0.99\linewidth]{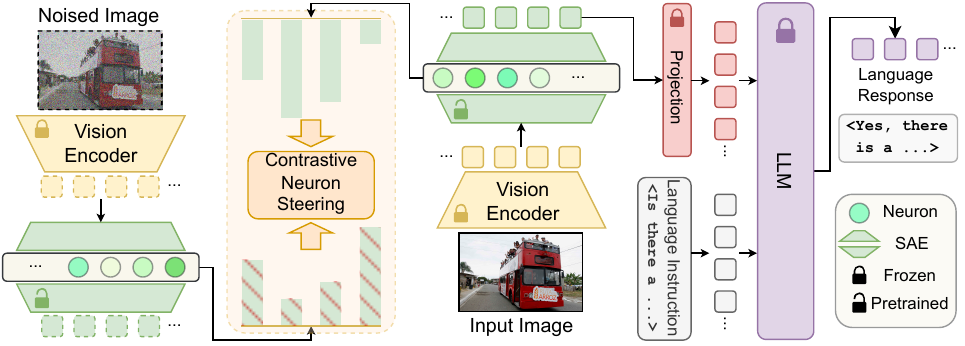}
    %\vspace{-10pt}
\caption{Framework overview. We integrate a pretrained SAE into the LVLM visual encoder to decompose dense visual features into sparse neurons, enabling neuron-level interpretation and control. We then apply Contrastive Neuron Steering (CNS) to identify image-specific neurons by comparing activations from clean and noisy inputs, selectively amplify these neurons, and produce more grounded visual representations, thereby reducing hallucinations.}
    %\vspace{-10pt}
    \label{fig:framework}
\end{figure}

\subsection{Neuron Intervention to Mitigate Hallucinations}

Building on the above analysis, we further investigate how neuron-level interventions influence model outputs and whether such interventions can be used to mitigate hallucinations. Our findings indicate that hallucinations are closely associated with disruptions in image-specific neurons, motivating direct manipulation of these neurons to improve visual grounding. Accordingly, we explore neuron-level intervention strategies that selectively amplify or suppress image-specific activations during inference.

We conduct controlled neuron interventions across a variety of visual scenarios, including concept insertion and removal, as well as neuron amplification and suppression. As illustrated in ~\Cref{fig:steering_two_neurons,fig:steering_dog_neurons,fig:steering_single_multi_insert,fig:steering_single_multi_supress}, adjusting neuron activations directly influences LVLM outputs: amplifying a neuron increases the likelihood that the corresponding visual concept appears in the model’s prediction, while suppressing a neuron reduces or eliminates that concept. These results demonstrate that neuron-level interventions provide a reliable and interpretable mechanism for systematically modulating model behavior.

We further illustrate how neuron-level intervention explains and mitigates hallucinations through concrete examples. In the black-apple case (\Cref{fig:case_black_apple}), a misleading neuron associated with red strawberries dominates the internal representation, suppressing apple-related neurons and leading the model to hallucinate the color “red.” By selectively suppressing this neuron or amplifying apple-specific neurons, the model correctly predicts the apple color as “black.” Similarly, in the sheep-and-dog example (\Cref{fig:case_sheep}), residual activation of dog-related neurons causes the model to hallucinate the presence of a dog; suppressing the corresponding neuron restores the correct prediction.

Overall, these results demonstrate that neuron-level intervention provides a fine-grained and interpretable mechanism for hallucination mitigation. By amplifying image-specific neurons that encode scene-relevant semantics and suppressing misleading activations, the model’s internal representations become better aligned with visual evidence. 

\section{Method}
\label{sec:method}

\subsection{Overview}

Building on our analysis, we propose \textbf{Contrastive Neuron Steering (CNS)} to mitigate hallucinations by selectively amplifying image-specific neurons that encode visually grounded semantics. CNS achieves this by comparing neuron activations between a noisy and the original clean image to identify image-specific neurons, then amplifying these neurons while suppressing non-informative activations to produce more robust and grounded visual embeddings. 

As illustrated in \Cref{fig:framework}, a pretrained SAE first decomposes dense visual features into sparse, interpretable neurons, which are then selectively enhanced via CNS. The enhanced visual features are fed into the LVLM at the prefilling stage, and they can be seamlessly combined with existing decoding-stage hallucination mitigation techniques.

\subsection{Contrastive Neuron Steering}
Given a clean image $v$ and its perturbed counterpart $v'$, we extract sparse neuron activations using a pretrained SAE, denoted as $z(v), z(v') \in \mathbb{R}^d$.
We compute a contrastive activation difference:
\begin{equation}
\Delta z = z(v) - z(v').
\end{equation}

\paragraph{Always-on Neuron Suppression (ANS).}
Always-on neurons typically exhibit large activation magnitudes and noticeable fluctuations, yet contribute little to image-specific semantics or final prediction quality.
Directly enhancing these neurons may introduce interference rather than improving visual grounding.
To address this issue, we propose {Always-on Neuron Suppression (ANS)}, which prevents always-on neurons from being steered and preserves only image-specific signals.

Concretely, we set the contrastive updates of always-on neurons to zero.
Let $\mathcal{A}$ denote the set of always-on neurons identified in Section~\ref{sec:analyse_sae}.
The $\Delta z$ is modified as:
\begin{equation}
\Delta z_i =
\begin{cases}
0, & i \in \mathcal{A}, \\
\Delta z_i, & \text{otherwise}.
\end{cases}
\end{equation}

Finally, CNS steers the clean activation by adding the filtered contrastive signal:
\begin{equation}
\tilde{z} = z(v) + \lambda \cdot \Delta z,
\end{equation}
where $\lambda$ controls the steering strength.
The enhanced activation $\tilde{z}$ is decoded into visual features.

\begin{table*}[t]
\vspace{-4pt}
    \caption{Results on the POPE (Accuracy and F1). $\uparrow$ indicates that higher is better.}
\vspace{-8pt}
    \label{tab:results_pope}
    \centering
    \resizebox{0.98\textwidth}{!}{
    \begin{tabular}{cl cc cc cc cc cc}
    \toprule
      \multirow{2}{*}{Setup} & \multirow{2}{*}{Method} 
     & \multicolumn{2}{c}{LLaVA-1.5} 
     & \multicolumn{2}{c}{InstructBLIP} 
     & \multicolumn{2}{c}{Qwen-VL}
     & \multicolumn{2}{c}{LLaVA-1.5-13B}
     & \multicolumn{2}{c}{LLaVA-Next} \\
    \arrayrulecolor{gray}
    \cmidrule(lr){3-4} \cmidrule(lr){5-6} \cmidrule(lr){7-8} \cmidrule(lr){9-10} \cmidrule(lr){11-12}
     &   & Accuracy $\uparrow$ & F1 $\uparrow$ 
       & Accuracy $\uparrow$ & F1 $\uparrow$ 
       & Accuracy $\uparrow$ & F1 $\uparrow$ 
       & Accuracy $\uparrow$ & F1 $\uparrow$ 
       & Accuracy $\uparrow$ & F1 $\uparrow$ \\
    \midrule

\multirow{4}{*}{Random}
& Vanilla      & 84.63 & 84.99 & 83.33 & 83.57 & 85.17 & 83.00 & 83.27 & 84.27 & 84.23 & 82.14 \\
& \cellcolor{gray!15} \quad + CNS & \cellcolor{gray!15}\textbf{87.76} & \cellcolor{gray!15}\textbf{87.89} 
               & \cellcolor{gray!15}\textbf{86.48} & \cellcolor{gray!15}\textbf{86.68} 
               & \cellcolor{gray!15}\textbf{88.23} & \cellcolor{gray!15}\textbf{86.29} 
               & \cellcolor{gray!15}\textbf{86.28} & \cellcolor{gray!15}\textbf{87.36} 
               & \cellcolor{gray!15}\textbf{87.18} & \cellcolor{gray!15}\textbf{85.62} \\
& VCD          & 84.57 & 85.02 & 84.60 & 84.49 & 84.69 & 82.91 & 83.47 & 84.52 & 83.48 & 81.36 \\
& \cellcolor{gray!15} \quad + CNS & \cellcolor{gray!15}\textbf{87.96} & \cellcolor{gray!15}\textbf{88.13} 
               & \cellcolor{gray!15}\textbf{87.82} & \cellcolor{gray!15}\textbf{87.97} 
               & \cellcolor{gray!15}\textbf{89.27} & \cellcolor{gray!15}\textbf{89.42} 
               & \cellcolor{gray!15}\textbf{86.62} & \cellcolor{gray!15}\textbf{87.83} 
               & \cellcolor{gray!15}\textbf{86.96} & \cellcolor{gray!15}\textbf{88.18} \\
\midrule

\multirow{4}{*}{Popular}
& Vanilla      & 81.33 & 82.33 & 76.00 & 77.94 & 84.50 & 82.50 & 80.57 & 82.19 & 82.33 & 80.44 \\
& \cellcolor{gray!15} \quad + CNS & \cellcolor{gray!15}\textbf{84.70} & \cellcolor{gray!15}\textbf{85.56} 
               & \cellcolor{gray!15}\textbf{79.07} & \cellcolor{gray!15}\textbf{80.71} 
               & \cellcolor{gray!15}\textbf{86.77} & \cellcolor{gray!15}\textbf{84.85} 
               & \cellcolor{gray!15}\textbf{83.42} & \cellcolor{gray!15}\textbf{85.01} 
               & \cellcolor{gray!15}\textbf{85.10} & \cellcolor{gray!15}\textbf{83.22} \\
& VCD          & 81.57 & 83.02 & 76.68 & 77.82 & 84.37 & 82.46 & 80.96 & 82.47 & 82.46 & 80.68 \\
& \cellcolor{gray!15} \quad + CNS & \cellcolor{gray!15}\textbf{83.53} & \cellcolor{gray!15}\textbf{84.73} 
               & \cellcolor{gray!15}\textbf{80.23} & \cellcolor{gray!15}\textbf{81.51} 
               & \cellcolor{gray!15}\textbf{88.17} & \cellcolor{gray!15}\textbf{86.78} 
               & \cellcolor{gray!15}\textbf{83.83} & \cellcolor{gray!15}\textbf{85.24} 
               & \cellcolor{gray!15}\textbf{85.21} & \cellcolor{gray!15}\textbf{83.94} \\
\midrule

\multirow{4}{*}{Adversarial}
& Vanilla      & 75.87 & 78.27 & 74.17 & 76.58 & 82.53 & 80.56 & 75.12 & 78.35 & 79.37 & 77.88 \\
& \cellcolor{gray!15} \quad + CNS & \cellcolor{gray!15}\textbf{79.16} & \cellcolor{gray!15}\textbf{81.52} 
               & \cellcolor{gray!15}\textbf{77.62} & \cellcolor{gray!15}\textbf{79.74} 
               & \cellcolor{gray!15}\textbf{85.86} & \cellcolor{gray!15}\textbf{83.92} 
               & \cellcolor{gray!15}\textbf{78.34} & \cellcolor{gray!15}\textbf{81.62} 
               & \cellcolor{gray!15}\textbf{82.57} & \cellcolor{gray!15}\textbf{81.23} \\
& VCD          & 76.13 & 78.62 & 74.62 & 76.82 & 82.79 & 80.82 & 75.64 & 78.68 & 79.72 & 78.23 \\
& \cellcolor{gray!15} \quad + CNS & \cellcolor{gray!15}\textbf{79.48} & \cellcolor{gray!15}\textbf{81.86} 
               & \cellcolor{gray!15}\textbf{77.92} & \cellcolor{gray!15}\textbf{79.96} 
               & \cellcolor{gray!15}\textbf{86.12} & \cellcolor{gray!15}\textbf{84.28} 
               & \cellcolor{gray!15}\textbf{78.82} & \cellcolor{gray!15}\textbf{82.08} 
               & \cellcolor{gray!15}\textbf{82.94} & \cellcolor{gray!15}\textbf{81.91} \\
\bottomrule
    \end{tabular}
    }
\end{table*}

\begin{table*}[t]
\vspace{-4pt}
    \caption{{Results on CHAIR (Max Token 128).} $\downarrow$ denotes lower is better.  -- denotes unavailable results.}
\vspace{-8pt}
    \label{tab:results_chair}
    \centering
    \resizebox{0.98\textwidth}{!}{
    \begin{tabular}{lcc cc cc cc cc}
        \toprule
        \multirow{2}{*}{Method} 
        & \multicolumn{2}{c}{LLaVA-1.5} 
        & \multicolumn{2}{c}{InstructBLIP} 
        & \multicolumn{2}{c}{Qwen-VL}
        & \multicolumn{2}{c}{LLaVA-1.5-13B}
        & \multicolumn{2}{c}{LLaVA-Next} \\
        \cmidrule(lr){2-3} \cmidrule(lr){4-5} \cmidrule(lr){6-7} \cmidrule(lr){8-9} \cmidrule(lr){10-11}
        & CHAIR$_S$ $\downarrow$ & CHAIR$_I$ $\downarrow$ 
        & CHAIR$_S$ $\downarrow$ & CHAIR$_I$ $\downarrow$ 
        & CHAIR$_S$ $\downarrow$ & CHAIR$_I$ $\downarrow$
        & CHAIR$_S$ $\downarrow$ & CHAIR$_I$ $\downarrow$
        & CHAIR$_S$ $\downarrow$ & CHAIR$_I$ $\downarrow$ \\
        \midrule

Vanilla   & 55.1 & 16.4 & 57.4 & 17.6 & 52.1 & 16.7 & 50.4 & 14.7 & 30.2 & 10.9 \\
\cellcolor{gray!15} \quad + CNS & 
\cellcolor{gray!15}\textbf{51.2} & \cellcolor{gray!15}\textbf{14.8} & 
\cellcolor{gray!15}\textbf{52.2} & \cellcolor{gray!15}\textbf{15.2} & 
\cellcolor{gray!15}\textbf{48.7} & \cellcolor{gray!15}\textbf{14.2} & 
\cellcolor{gray!15}\textbf{47.6} & \cellcolor{gray!15}\textbf{13.4} & 
\cellcolor{gray!15}\textbf{29.2} & \cellcolor{gray!15}\textbf{9.6} \\
\midrule
 PAI  & 53.1 & 15.1 & 54.2 & 15.6 & 49.4 & 15.6  & — & — & — & — \\
 ICD  & 54.6 & 15.4 & 55.2 & 16.2 & 49.8 & 16.2  & — & — & — & — \\
VCD       & 53.4 & 15.8 & 55.1 & 15.7 & 49.1 & 15.6 & 48.6 & 14.1 & 29.4 & 10.2 \\
\cellcolor{gray!15} \quad + CNS & 
\cellcolor{gray!15}\textbf{50.1} & \cellcolor{gray!15}\textbf{13.7} & 
\cellcolor{gray!15}\textbf{52.2} & \cellcolor{gray!15}\textbf{14.2} & 
\cellcolor{gray!15}\textbf{47.8} & \cellcolor{gray!15}\textbf{13.6} & 
\cellcolor{gray!15}\textbf{47.3} & \cellcolor{gray!15}\textbf{13.2} & 
\cellcolor{gray!15}\textbf{27.9} & \cellcolor{gray!15}\textbf{9.1} \\
M3ID    & 56.6 & 15.8 & 62.3 & 18.2 & 49.8 & 17.4 & — & — & — & — \\
\cellcolor{gray!15} \quad + CNS & 
\cellcolor{gray!15}\textbf{52.4} & \cellcolor{gray!15}\textbf{14.2} & 
\cellcolor{gray!15}\textbf{56.3} & \cellcolor{gray!15}\textbf{15.4} & 
\cellcolor{gray!15}\textbf{48.2} & \cellcolor{gray!15}\textbf{15.4} & 
 — & — & — & — \\
\midrule
 
Woodpecker& 57.6 & 16.7 & 60.8 & 17.6 & 51.8 & 16.3 & — & — & — & — \\
HALC      & 51.0 & 14.8 & 53.8 & 15.7 & 49.6 & 15.4 & — & — & — & — \\
\cellcolor{gray!15} \quad + CNS & 
\cellcolor{gray!15}\textbf{49.6} & \cellcolor{gray!15}\textbf{14.1} & 
\cellcolor{gray!15}\textbf{51.2} & \cellcolor{gray!15}\textbf{14.6} & 
\cellcolor{gray!15}\textbf{48.8} & \cellcolor{gray!15}\textbf{14.2} & 
 — & — & — & — \\
AGLA      & 52.4 & 14.6 & 54.8 & 16.2 & 49.8 & 15.6 & — & — & — & — \\
\cellcolor{gray!15} \quad + CNS & 
\cellcolor{gray!15}\textbf{50.4} & \cellcolor{gray!15}\textbf{13.6} & 
\cellcolor{gray!15}\textbf{51.8} & \cellcolor{gray!15}\textbf{15.3} & 
\cellcolor{gray!15}\textbf{48.6} & \cellcolor{gray!15}\textbf{14.3} & 
 — & — & — & — \\
\midrule

OPERA     & 51.6 & 14.2 & 54.2 & 14.8 & 48.6 & 14.6    & — & — & — & — \\
VAF  & 50.1 & 14.2 & 53.4 & 15.1 & 48.7 & 14.4  & — & — & — & — \\
ONLY      & 49.8 & 14.3 & 52.2 & 15.5 & 48.0 & 14.3 & — & — & — & — \\
\cellcolor{gray!15} \quad + CNS & 
\cellcolor{gray!15}\textbf{48.4} & \cellcolor{gray!15}\textbf{13.2} & 
\cellcolor{gray!15}\textbf{51.4} & \cellcolor{gray!15}\textbf{14.1} & 
\cellcolor{gray!15}\textbf{47.2} & \cellcolor{gray!15}\textbf{13.2} & 
— & — & — & — \\
VAR      & 49.6 & 14.1 & 52.3 & 14.7 & 48.4 & 13.9  & — & — & — & — \\
\cellcolor{gray!15} \quad + CNS & 
\cellcolor{gray!15}\textbf{48.2} & \cellcolor{gray!15}\textbf{12.6} & 
\cellcolor{gray!15}\textbf{50.9} & \cellcolor{gray!15}\textbf{13.5} & 
\cellcolor{gray!15}\textbf{46.9} & \cellcolor{gray!15}\textbf{12.3} & 
— & — & — & — \\

        \bottomrule
    \end{tabular}
    }
\end{table*}

\section{Experiments}
\label{sec:experiment}

\noindent \textbf{Benchmarks.}  
To comprehensively evaluate our method, we conduct experiments on two categories of benchmarks. 
(1) \emph{Hallucination-focused benchmarks}: POPE~\cite{pope}, CHAIR~\cite{chair}, HallusionBench~\cite{hallusionbench}, and AMBER~\cite{amber}, which evaluate hallucinations in classification, captioning, visual consistency, and visually grounded reasoning.
(2) \emph{General-purpose benchmarks}: VizWiz~\cite{gurari2018vizwiz}, MME~\cite{mme}, LLaVA-Wild~\cite{llava}, and MM-Vet~\cite{yu2023mmvet}, covering a wide range of visual understanding, reasoning, and real-world multimodal tasks.

\noindent \textbf{Evaluated LVLMs.}
We evaluate our method on several representative open-source LVLMs, including {LLaVA-1.5}~\cite{liu2024llava1.5}, {InstructBLIP}~\cite{instructblip}, and {Qwen-VL}~\cite{qwenvl}.
We additionally consider a larger scale variant, {LLaVA-1.5-13B}, as well as the stronger and newer {LLaVA-NeXT~\cite{llavanext}}.
Together, these models cover diverse backbone architectures and model scales, providing a comprehensive evaluation setting.
Following prior works~\cite{vcd,only}, we apply sampling-based decoding in default. Unless otherwise specified, {LLaVA-1.5} is used as the default model.

\noindent \textbf{Baselines.}
We compare CNS with various hallucination mitigation methods:
(1) \emph{contrastive decoding} methods (VCD~\cite{vcd}, PAI~\cite{pai}, M3ID~\cite{m3id}, ICD~\cite{icd}); 
(2) \emph{auxiliary expert model} methods (HALC~\cite{halc}, AGLA~\cite{agla}, Woodpecker~\cite{woodpecker}); and 
(3) \emph{static internal signal} methods (OPERA~\cite{opera}, VAF~\cite{yin2025clearsight_vaf}, VAR~\cite{kang2025see_attention_sinks}, ONLY~\cite{only}).
Unless otherwise specified, we adopt sampling-based decoding as the default.

\begin{table}[t]
\centering
\vspace{-8pt}
\caption{Results on the HallusionBench.}
\vspace{-8pt}
\label{tab:results_HallusionBench} 
\resizebox{0.98\linewidth}{!}{
\begin{tabular}{lcccc}
\toprule
Methods & fACC $\uparrow$ & qACC $\uparrow$ & ${easy}$A $\uparrow$ & ${hard}$A $\uparrow$ \\ 
\midrule
Vanilla   & 17.9 & 8.13 & 36.0 & 36.7 \\
\cellcolor{gray!15} \quad + CNS & \cellcolor{gray!15}\textbf{18.2} 
            & \cellcolor{gray!15}\textbf{8.39} 
            & \cellcolor{gray!15}\textbf{36.4} 
            & \cellcolor{gray!15}\textbf{37.4} \\
VCD       & 15.6 & 8.47 & 34.8 & 35.2 \\
\cellcolor{gray!15} \quad + CNS & \cellcolor{gray!15}\textbf{18.7} 
            & \cellcolor{gray!15}\textbf{9.12} 
            & \cellcolor{gray!15}\textbf{36.8} 
            & \cellcolor{gray!15}\textbf{37.8} \\
\bottomrule
\end{tabular}
}
\end{table}

\begin{table}[t]
\vspace{-8pt}
\caption{Results on AMBER Generative Subset.}
\vspace{-8pt}
\label{tab:results_amber}
\centering
\resizebox{0.98\linewidth}{!}{
\begin{tabular}{lcccc}
\toprule
\textbf{Method} & \textbf{CHAIR ($\downarrow$)} & \textbf{Cover ($\uparrow$)} & \textbf{Hall ($\downarrow$)} & \textbf{Cog ($\downarrow$)} \\
\midrule
Vanilla & 7.8 & 51.0 & 36.4 & 4.2 \\
\cellcolor{gray!15} \quad + CNS & \cellcolor{gray!15}\textbf{7.1} 
            & \cellcolor{gray!15}\textbf{52.4} 
            & \cellcolor{gray!15}\textbf{34.7} 
            & \cellcolor{gray!15}\textbf{3.9} \\
VCD & 7.4 & 50.8 & 36.1 & 4.0 \\
\cellcolor{gray!15} \quad + CNS & \cellcolor{gray!15}\textbf{6.9} 
            & \cellcolor{gray!15}\textbf{52.8} 
            & \cellcolor{gray!15}\textbf{34.2} 
            & \cellcolor{gray!15}\textbf{3.7} \\
\bottomrule
\end{tabular}
}
\vspace{-8pt}

\end{table}

\begin{table}[t]
\centering
\vspace{-4pt}
\caption{Results on multiple general vision-language benchmarks.}
\vspace{-8pt}
\label{tab:results_general_benchmarks}
\resizebox{1\linewidth}{!}{
\begin{tabular}{l c c c c}
\toprule
\multirow{2}{*}{Methods} 
& VizWiz 
& MME
& LLaVA-Wild 
& MM-Vet \\
\cmidrule(lr){2-2} 
\cmidrule(lr){3-3} 
\cmidrule(lr){4-4} 
\cmidrule(lr){5-5} 
& Accuracy $\uparrow$ 
& Overall $\uparrow$
& Average $\uparrow$
& Total $\uparrow$ \\
\midrule
Vanilla      & 50.00 & 1864.68 & 64.80 & 31.1 \\
\cellcolor{gray!15} \quad + CNS  & \cellcolor{gray!15}\textbf{50.92} 
             & \cellcolor{gray!15}\textbf{1883.57} 
             & \cellcolor{gray!15}\textbf{65.92} 
             & \cellcolor{gray!15}\textbf{32.4} \\
VCD          & 45.62 & 1873.82 & 63.28 & 30.6 \\
\cellcolor{gray!15} \quad + CNS  & \cellcolor{gray!15}\textbf{51.82} 
             & \cellcolor{gray!15}\textbf{1892.68} 
             & \cellcolor{gray!15}\textbf{66.84} 
             & \cellcolor{gray!15}\textbf{32.8} \\
\bottomrule
\end{tabular}
}
\vspace{-12pt}

\end{table}

\begin{table}[t]
\centering
\vspace{-8pt}
\caption{Inference efficiency.}
\vspace{-8pt}
\label{tab:efficiency}
\resizebox{0.85\linewidth}{!}{
\begin{tabular}{lcc}
\toprule
Method & Avg. Latency $\downarrow$ & GPU Memory $\downarrow$ \\
\midrule
Vanilla & 3.19 s ($\times$1.00) & 14945 MB ($\times$1.00) \\
VCD & 6.44 s ($\times$2.02) & 15749 MB ($\times$1.05) \\
M3ID & 6.52 s ($\times$2.04) & 15575 MB ($\times$1.04) \\
OPERA & 22.64 s ($\times$7.10) & 22706 MB ($\times$1.52) \\
Woodpecker & 9.76 s ($\times$3.06) & 22199 MB ($\times$1.49) \\
HALC & 20.64 s ($\times$6.47) & 23084 MB ($\times$1.54) \\
\textbf{CNS (Ours)} & \textbf{3.64 s ($\times$1.14)} & \textbf{15226 MB ($\times$1.02)} \\
\bottomrule
\end{tabular}}
\vspace{-4pt}

\end{table}

\textbf{Implementation Details.}  
To obtain interpretable and sparse visual representations for neuron-level analysis, we train a Matryoshka BatchTopK SAE~\cite{bussmann2024matryoshka} on ImageNet~\cite{deng2009imagenet} using image features extracted from each LVLM's visual encoder. The Matryoshka groups are set as $\mathcal{M}=\{0.0625\omega,0.1875\omega,0.4375\omega,\omega\}$ to progressively increase the number of active neurons per level. We fix the maximum number of non-zero latent neurons to $K=20$ and set the expansion factor to $64$. Training is performed for $10^5$ steps with a batch size of 4096 using Adam~\cite{adam} with a learning rate of $\frac{16}{125\sqrt{\omega}}$. This setup balances sparsity and reconstruction quality, yielding robust neuron-level features for subsequent analysis and enhancement. All experiments are conducted on a single NVIDIA RTX 6000 GPU (48GB).

\subsection{Main Experimental Results}

\noindent \textbf{Results on Hallucination Benchmarks.}  
As shown in \Cref{tab:results_pope,tab:results_chair,tab:results_amber,tab:results_HallusionBench}, CNS consistently outperforms existing decoding-stage hallucination mitigation methods across model architectures and scales. By operating at the representation level, CNS identifies and selectively enhances image-specific neurons while suppressing non-informative or noisy activations. This removes hallucination-inducing factors from the input and produces cleaner, semantically grounded visual features, resulting in more accurate and reliable model outputs. Furthermore, CNS can be seamlessly combined with existing decoding methods, leading to additional performance gains.

\noindent \textbf{Results on General Benchmarks.}  
As shown in \Cref{tab:results_general_benchmarks}, CNS maintains and in some cases slightly improves performance on general multimodal benchmarks. Rather than simply constraining output distributions, CNS refines the internal visual representations, enhancing informative neurons and reducing noise. This allows the model to better leverage its inherent multimodal reasoning capabilities, mitigating hallucinations while preserving and in some cases strengthening general visual understanding and reasoning.

\noindent \textbf{Efficiency Comparison.}
As shown in \Cref{tab:efficiency}, we evaluate inference efficiency on the CHAIR benchmark using LLaVA-7B with a single NVIDIA A6000 GPU.
Compared to multi-pass decoding-stage methods or approaches that rely on additional expert models, which incur substantial overhead due to repeated decoding or extra model forward passes, CNS introduces only minimal overhead.
It operates exclusively at the prefilling stage, requiring a single additional forward pass through the visual encoder, while the lightweight SAE with two linear layers adds negligible computational cost.
Overall, CNS enhances visual representation quality, effectively reduces hallucinations, and maintains high computational efficiency.

\begin{figure}[t]
    \centering
    \includegraphics[width=0.95\linewidth]{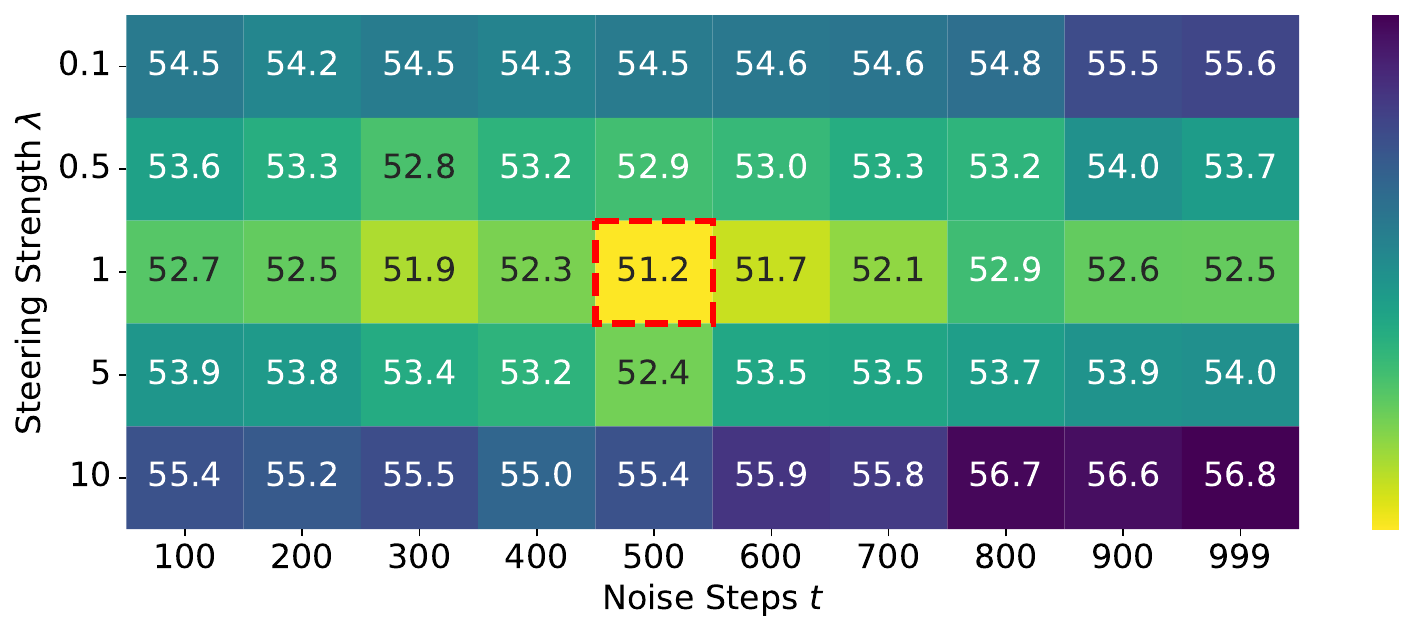}
\vspace{-8pt}
\caption{Effect of noise steps and steering strength $\lambda$. }
\vspace{-8pt}
    
    \label{fig:abla_step_lambda}
\end{figure}

\subsection{Ablation Studies}
\label{sec:ablation}

\noindent \textbf{Effect of ANS.}  
To evaluate the role of ANS, we compare the full CNS with a variant that does not suppress contrastive updates on always-on neurons (w/o ANS) across different models and scales. As shown in~\Cref{tab:abla_ans}, removing ANS leads to a consistent but moderate performance drop on both CHAIR$_S$ and CHAIR$_I$. This result indicates that suppressing always-on neurons helps reduce interference from non-informative activations, allowing CNS to better focus enhancement on image-specific neurons and thereby produce more reliable visual representations that effectively mitigate hallucinations.

\noindent \textbf{Noise Step $t$.}  
As shown in~\Cref{fig:abla_step_lambda}, we investigate the influence of the number of noise steps used to generate perturbed inputs on LLaVA-1.5 7b. Performance initially improves as the noise step increases, because stronger perturbations more effectively reveal neurons that are sensitive to input variations, allowing CNS to better identify image-specific neurons for enhancement. Beyond roughly 500 steps, further increases slightly degrade performance, as excessively strong perturbations can suppress even stable image-specific neurons and introduce noisy contrastive signals, indicating an optimal value around $t=500$.

\noindent \textbf{Steering Strength $\lambda$.}  
As shown in~\Cref{fig:abla_step_lambda}, we study the effect of the neuron enhancement strength $\lambda$ on LLaVA-1.5 7b.  The best overall results are observed at $\lambda=1$. Moderate values sufficiently amplify image-specific neurons without causing instability. Smaller values (e.g., 0.1) result in limited enhancement, while excessively large values may amplify spurious activations and reduce performance.

\textbf{For more analysis, experiments, and visualizations, please refer to the appendix.}

\begin{table}[t]
\vspace{-4pt}
\caption{Ablation study on the ANS component within CNS.}
\vspace{-8pt}
\label{tab:abla_ans}
\centering
\resizebox{0.9\linewidth}{!}{
\begin{tabular}{l lcc}
\toprule
\textbf{Base Model} & \textbf{Method} & \textbf{CHAIR$_S$} $\downarrow$ & \textbf{CHAIR$_I$} $\downarrow$ \\
\midrule
\multirow{3}{*}{LLaVA-1.5-7B}
 & Vanilla           & 55.1 & 16.4 \\
 & \quad + CNS             & \textbf{51.2} & \textbf{14.8} \\
 & \quad + CNS w/o ANS     &  52.4 & 15.3 \\
\midrule
\multirow{3}{*}{LLaVA-1.5-13B}
 & Vanilla           & 50.4 & 14.7 \\
 & \quad + CNS             & \textbf{47.6} & \textbf{13.4} \\
 & \quad + CNS w/o ANS     & 48.5 & 13.9 \\
\midrule
\multirow{3}{*}{LLaVA-Next}
 & Vanilla           & 30.2 & 10.9 \\
 & \quad + CNS             & \textbf{29.2} & \textbf{9.6} \\
 & \quad + CNS w/o ANS     & 29.7 & 10.2 \\
\midrule
\multirow{3}{*}{Qwen-VL}
 & Vanilla           & 52.1 & 16.7 \\
 & \quad + CNS             & \textbf{48.7} & \textbf{14.2} \\
 & \quad + CNS w/o ANS     & 49.8 & 14.9 \\
\midrule
\multirow{3}{*}{InstructBLIP}
 & Vanilla           & 57.4 & 17.6 \\
 & \quad + CNS             & \textbf{52.2} & \textbf{15.2} \\
 & \quad + CNS w/o ANS     & 53.6 & 15.9 \\
\bottomrule
\end{tabular}
}
\vspace{-12pt}
\end{table}

\section{Conclusion}
\label{sec:conclusion}

We present a representation-level framework for understanding and mitigating hallucinations in LVLMs. By decomposing visual embeddings into sparse interpretable neurons, we find that hallucinations arise from spurious or disrupted image-specific neurons and show that neuron-level interventions can correct outputs. Our CNS selectively enhances image-specific neurons via contrastive analysis of clean and noisy inputs, improving visual grounding, reducing hallucinations, and remaining fully compatible with existing decoding-stage methods.

\clearpage
\clearpage

\section{Impact Statement}
This paper presents work whose goal is to advance the field of Machine Learning. There are many potential societal consequences of our work, none which we feel must be specifically highlighted here.
% \input{tabs/chair}
% \input{tabs/pope}
% \input{tabs/ablation}

% \clearpage

\bibliography{main}

@article{zhu2023minigpt4,
	title        = {MiniGPT-4: Enhancing Vision-Language Understanding with Advanced Large Language Models},
	author       = {Deyao Zhu and Jun Chen and Xiaoqian Shen and Xiang Li and Mohamed Elhoseiny},
	year         = 2023,
	journal      = {arXiv preprint arXiv:2304.10592}
}

@article{pope,
	title        = {Evaluating Object Hallucination in Large Vision-Language Models},
	author       = {Yifan Li and Yifan Du and Kun Zhou and Jinpeng Wang and Wayne Xin Zhao and Ji-Rong Wen},
	year         = 2023,
	journal      = {arXiv preprint arXiv:2305.10355}
}

@article{liu2023mitigating,
	title        = {Mitigating Hallucination in Large Multi-Modal Models via Robust Instruction Tuning},
	author       = {Fuxiao Liu and Kevin Lin and Linjie Li and Jianfeng Wang and Yaser Yacoob and Lijuan Wang},
	year         = 2023,
	journal      = {arXiv preprint arXiv:2306.14565}
}

@article{monosamantic,
	title        = {Sparse autoencoders learn monosemantic features in vision-language models},
	author       = {Pach, Mateusz and Karthik, Shyamgopal and Bouniot, Quentin and Belongie, Serge and Akata, Zeynep},
	year         = 2025,
	journal      = {arXiv preprint arXiv:2504.02821}
}

@article{instructblip,
	title        = {Instructblip: Towards general-purpose vision-language models with instruction tuning},
	author       = {Dai, Wenliang and Li, Junnan and Li, Dongxu and Tiong, Anthony and Zhao, Junqi and Wang, Weisheng and Li, Boyang and Fung, Pascale N and Hoi, Steven},
	year         = 2023,
	journal      = {NIPS},
	volume       = 36,
	pages        = {49250--49267}
}

@article{qwenvl,
	title        = {Qwen technical report},
	author       = {Bai, Jinze and Bai, Shuai and Chu, Yunfei and Cui, Zeyu and Dang, Kai and Deng, Xiaodong and Fan, Yang and Ge, Wenbin and Han, Yu and Huang, Fei and others},
	year         = 2023,
	journal      = {arXiv preprint arXiv:2309.16609}
}

@inproceedings{gunjal2024detecting,
  title={Detecting and preventing hallucinations in large vision language models},
  author={Gunjal, Anisha and Yin, Jihan and Bas, Erhan},
  booktitle={AAAI},
  number={16},
  year={2024}
}

@inproceedings{zhibo2023overcoming,
	title        = {Overcoming Language Priors with Counterfactual Inference for Visual Question Answering},
	author       = {Zhibo, Ren and Huizhen, Wang and Muhua, Zhu and Yichao, Wang and Tong, Xiao and Jingbo, Zhu},
	year         = 2023,
	booktitle    = {Proceedings of the 22nd Chinese National Conference on Computational Linguistics},
	pages        = {600--610}
}

@article{han2022visual,
	title        = {Visual Perturbation-aware Collaborative Learning for Overcoming the Language Prior Problem},
	author       = {Han, Yudong and Nie, Liqiang and Yin, Jianhua and Wu, Jianlong and Yan, Yan},
	year         = 2022,
	journal      = {arXiv preprint arXiv:2207.11850}
}

@article{agrawal2016analyzing,
	title        = {Analyzing the behavior of visual question answering models},
	author       = {Agrawal, Aishwarya and Batra, Dhruv and Parikh, Devi},
	year         = 2016,
	journal      = {arXiv preprint arXiv:1606.07356}
}

@inproceedings{agarwal2020towards,
	title        = {Towards causal vqa: Revealing and reducing spurious correlations by invariant and covariant semantic editing},
	author       = {Agarwal, Vedika and Shetty, Rakshith and Fritz, Mario},
	year         = 2020,
	booktitle    = {CVPR},
	pages        = {9690--9698}
}

@article{sun2023aligning,
	title        = {Aligning large multimodal models with factually augmented rlhf},
	author       = {Sun, Zhiqing and Shen, Sheng and Cao, Shengcao and Liu, Haotian and Li, Chunyuan and Shen, Yikang and Gan, Chuang and Gui, Liang-Yan and Wang, Yu-Xiong and Yang, Yiming and others},
	year         = 2023,
	journal      = {arXiv preprint arXiv:2309.14525}
}

@article{lee2018hallucinations,
	title        = {Hallucinations in neural machine translation},
	author       = {Lee, Katherine and Firat, Orhan and Agarwal, Ashish and Fannjiang, Clara and Sussillo, David},
	year         = 2018,
	journal      = {OpenReview}
}

@article{rajamanoharan2024jumping,
	title        = {Jumping ahead: Improving reconstruction fidelity with jumprelu sparse autoencoders},
	author       = {Rajamanoharan, Senthooran and Lieberum, Tom and Sonnerat, Nicolas and Conmy, Arthur and Varma, Vikrant and Kram{\'a}r, J{\'a}nos and Nanda, Neel},
	year         = 2024,
	journal      = {arXiv preprint arXiv:2407.14435}
}

@article{mme,
	title        = {MME: A Comprehensive Evaluation Benchmark for Multimodal Large Language Models},
	author       = {Fu, Chaoyou and Chen, Peixian and Shen, Yunhang and Qin, Yulei and Zhang, Mengdan and Lin, Xu and Qiu, Zhenyu and Lin, Wei and Yang, Jinrui and Zheng, Xiawu and others},
	year         = 2023,
	journal      = {arXiv preprint arXiv:2306.13394}
}

@inproceedings{m3id,
	title        = {Multi-modal hallucination control by visual information grounding},
	author       = {Favero, Alessandro and Zancato, Luca and Trager, Matthew and Choudhary, Siddharth and Perera, Pramuditha and Achille, Alessandro and Swaminathan, Ashwin and Soatto, Stefano},
	year         = 2024,
	booktitle    = {CVPR},
	pages        = {14303--14312}
}

@article{woodpecker,
	title        = {Woodpecker: Hallucination Correction for Multimodal Large Language Models},
	author       = {Yin, Shukang and Fu, Chaoyou and Zhao, Sirui and Xu, Tong and Wang, Hao and Sui, Dianbo and Shen, Yunhang and Li, Ke and Sun, Xing and Chen, Enhong},
	year         = 2023,
	journal      = {arXiv preprint arXiv:2310.16045}
}

@inproceedings{vcd,
	title        = {Mitigating object hallucinations in large vision-language models through visual contrastive decoding},
	author       = {Leng, Sicong and Zhang, Hang and Chen, Guanzheng and Li, Xin and Lu, Shijian and Miao, Chunyan and Bing, Lidong},
	year         = 2024,
	booktitle    = {CVPR},
	pages        = {13872--13882}
}

@article{halc,
	title        = {Halc: Object hallucination reduction via adaptive focal-contrast decoding},
	author       = {Chen, Zhaorun and Zhao, Zhuokai and Luo, Hongyin and Yao, Huaxiu and Li, Bo and Zhou, Jiawei},
	year         = 2024,
	journal      = {arXiv preprint arXiv:2403.00425}
}

@article{liu2024survey,
	title        = {A survey on hallucination in large vision-language models},
	author       = {Liu, Hanchao and Xue, Wenyuan and Chen, Yifei and Chen, Dapeng and Zhao, Xiutian and Wang, Ke and Hou, Liping and Li, Rongjun and Peng, Wei},
	year         = 2024,
	journal      = {arXiv preprint arXiv:2402.00253}
}

@article{zhu2024ibd,
	title        = {IBD: Alleviating Hallucinations in Large Vision-Language Models via Image-Biased Decoding},
	author       = {Zhu, Lanyun and Ji, Deyi and Chen, Tianrun and Xu, Peng and Ye, Jieping and Liu, Jun},
	year         = 2024,
	journal      = {arXiv preprint arXiv:2402.18476}
}

@article{lee2023volcano,
	title        = {Volcano: mitigating multimodal hallucination through self-feedback guided revision},
	author       = {Lee, Seongyun and Park, Sue Hyun and Jo, Yongrae and Seo, Minjoon},
	year         = 2023,
	journal      = {arXiv preprint arXiv:2311.07362}
}

@article{opera,
	title        = {Opera: Alleviating hallucination in multi-modal large language models via over-trust penalty and retrospection-allocation},
	author       = {Huang, Qidong and Dong, Xiaoyi and Zhang, Pan and Wang, Bin and He, Conghui and Wang, Jiaqi and Lin, Dahua and Zhang, Weiming and Yu, Nenghai},
	year         = 2023,
	journal      = {arXiv preprint arXiv:2311.17911}
}

@article{only,
	title        = {ONLY: One-Layer Intervention Sufficiently Mitigates Hallucinations in Large Vision-Language Models},
	author       = {Wan, Zifu and Zhang, Ce and Yong, Silong and Ma, Martin Q and Stepputtis, Simon and Morency, Louis-Philippe and Ramanan, Deva and Sycara, Katia and Xie, Yaqi},
	year         = 2025,
	journal      = {arXiv preprint arXiv:2507.00898}
}

@article{yue2024less,
	title        = {Less is More: Mitigating Multimodal Hallucination from an EOS Decision Perspective},
	author       = {Yue, Zihao and Zhang, Liang and Jin, Qin},
	year         = 2024,
	journal      = {arXiv preprint arXiv:2402.14545}
}

@article{chair,
	title        = {Object hallucination in image captioning},
	author       = {Rohrbach, Anna and Hendricks, Lisa Anne and Burns, Kaylee and Darrell, Trevor and Saenko, Kate},
	year         = 2018,
	journal      = {arXiv preprint arXiv:1809.02156}
}

@article{zhou2024aligning,
	title        = {Aligning Modalities in Vision Large Language Models via Preference Fine-tuning},
	author       = {Zhou, Yiyang and Cui, Chenhang and Rafailov, Rafael and Finn, Chelsea and Yao, Huaxiu},
	year         = 2024,
	journal      = {arXiv preprint arXiv:2402.11411}
}

@article{dola,
	title        = {Dola: Decoding by contrasting layers improves factuality in large language models},
	author       = {Chuang, Yung-Sung and Xie, Yujia and Luo, Hongyin and Kim, Yoon and Glass, James and He, Pengcheng},
	year         = 2023,
	journal      = {arXiv preprint arXiv:2309.03883}
}

@article{bussmann2024batchtopk,
	title        = {Batchtopk sparse autoencoders},
	author       = {Bussmann, Bart and Leask, Patrick and Nanda, Neel},
	year         = 2024,
	journal      = {arXiv preprint arXiv:2412.06410}
}

@article{makhzani2013k,
	title        = {K-sparse autoencoders},
	author       = {Makhzani, Alireza and Frey, Brendan},
	year         = 2013,
	journal      = {arXiv preprint arXiv:1312.5663}
}

@article{bricken2023towards,
	title        = {Towards monosemanticity: Decomposing language models with dictionary learning},
	author       = {Bricken, Trenton and Templeton, Adly and Batson, Joshua and Chen, Brian and Jermyn, Adam and Conerly, Tom and Turner, Nick and Anil, Cem and Denison, Carson and Askell, Amanda and others},
	year         = 2023,
	journal      = {Transformer Circuits Thread},
	volume       = 2
}

@article{bussmann2024matryoshka,
	title        = {Learning Multi-Level Features with Matryoshka SAEs},
	author       = {Bussmann, Bart and Leask, Patrick and Nanda, Neel},
	year         = 2024,
	journal      = {AI Alignment Forum}
}

@article{nabeshima2024matryoshka,
	title        = {Matryoshka Sparse Autoencoders},
	author       = {Nabeshima, Noa},
	year         = 2024,
	journal      = {AI Alignment Forum}
}

@article{olshausen1997sparse,
	title        = {Sparse coding with an overcomplete basis set: A strategy employed by V1?},
	author       = {Olshausen, Bruno A and Field, David J},
	year         = 1997,
	journal      = {Vision research},
	publisher    = {Elsevier},
	volume       = 37,
	number       = 23,
	pages        = {3311--3325}
}

@article{templeton2024scaling,
	title        = {Scaling Monosemanticity: Extracting Interpretable Features from Claude 3 Sonnet},
	author       = {Templeton, Adly and Conerly, Tom and Marcus, Jonathan and Lindsey, Jack and Bricken, Trenton and Chen, Brian and Pearce, Adam and Citro, Craig and Ameisen, Emmanuel and Jones, Andy and Cunningham, Hoagy and Turner, Nicholas L and McDougall, Callum and MacDiarmid, Monte and Freeman, C. Daniel and Sumers, Theodore R. and Rees, Edward and Batson, Joshua and Jermyn, Adam and Carter, Shan and Olah, Chris and Henighan, Tom},
	year         = 2024,
	journal      = {Transformer Circuits Thread},
}

@misc{durmus2024steering,
	title        = {Evaluating Feature Steering: A Case Study in Mitigating Social Biases},
	author       = {Esin Durmus and Alex Tamkin and Jack Clark and Jerry Wei and Jonathan Marcus and Joshua Batson and Kunal Handa and Liane Lovitt and Meg Tong and Miles McCain and Oliver Rausch and Saffron Huang and Sam Bowman and Stuart Ritchie and Tom Henighan and Deep Ganguli},
	year         = 2024,
	date         = {2024-10-25}
}

@article{thasarathan2025universal,
	title        = {Universal Sparse Autoencoders: Interpretable Cross-Model Concept Alignment},
	author       = {Thasarathan, Harrish and Forsyth, Julian and Fel, Thomas and Kowal, Matthew and Derpanis, Konstantinos},
	year         = 2025,
	journal      = {arXiv preprint arXiv:2502.03714}
}

@article{Revelio,
	title        = {Revelio: Interpreting and leveraging semantic information in diffusion models},
	author       = {Kim, Dahye and Thomas, Xavier and Ghadiyaram, Deepti},
	year         = 2024,
	journal      = {arXiv preprint arXiv:2411.16725}
}

@inproceedings{deng2009imagenet,
	title        = {Imagenet: A large-scale hierarchical image database},
	author       = {Deng, Jia and Dong, Wei and Socher, Richard and Li, Li-Jia and Li, Kai and Fei-Fei, Li},
	year         = 2009,
	booktitle    = {2009 IEEE conference on computer vision and pattern recognition},
	pages        = {248--255},
	organization = {Ieee}
}

@misc{adam,
	title        = {Adam: A Method for Stochastic Optimization},
	author       = {Diederik P. Kingma and Jimmy Ba},
	year         = 2017,
}

@inproceedings{cd,
	title        = {Contrastive Decoding: Open-ended Text Generation as Optimization},
	author       = {Li, Xiang Lisa and Holtzman, Ari and Fried, Daniel and Liang, Percy and Eisner, Jason and Hashimoto, Tatsunori B and Zettlemoyer, Luke and Lewis, Mike},
	year         = 2023,
	booktitle    = {ACL},
	pages        = {12286--12312}
}

@misc{openai2024gpt4technicalreport,
	title        = {GPT-4 Technical Report},
	author       = {OpenAI and Josh Achiam and Steven Adler and Sandhini Agarwal and Lama Ahmad and Ilge Akkaya and Florencia Leoni Aleman and Diogo Almeida and Janko Altenschmidt and Sam Altman and Shyamal Anadkat and Red Avila and Igor Babuschkin and Suchir Balaji and Valerie Balcom and Paul Baltescu and Haiming Bao and Mohammad Bavarian and Jeff Belgum and Irwan Bello and Jake Berdine and Gabriel Bernadett-Shapiro and Christopher Berner and Lenny Bogdonoff and Oleg Boiko and Madelaine Boyd and Anna-Luisa Brakman and Greg Brockman and Tim Brooks and Miles Brundage and Kevin Button and Trevor Cai and Rosie Campbell and Andrew Cann and Brittany Carey and Chelsea Carlson and Rory Carmichael and Brooke Chan and Che Chang and Fotis Chantzis and Derek Chen and Sully Chen and Ruby Chen and Jason Chen and Mark Chen and Ben Chess and Chester Cho and Casey Chu and Hyung Won Chung and Dave Cummings and Jeremiah Currier and Yunxing Dai and Cory Decareaux and Thomas Degry and Noah Deutsch and Damien Deville and Arka Dhar and David Dohan and Steve Dowling and Sheila Dunning and Adrien Ecoffet and Atty Eleti and Tyna Eloundou and David Farhi and Liam Fedus and Niko Felix and Simón Posada Fishman and Juston Forte and Isabella Fulford and Leo Gao and Elie Georges and Christian Gibson and Vik Goel and Tarun Gogineni and Gabriel Goh and Rapha Gontijo-Lopes and Jonathan Gordon and Morgan Grafstein and Scott Gray and Ryan Greene and Joshua Gross and Shixiang Shane Gu and Yufei Guo and Chris Hallacy and Jesse Han and Jeff Harris and Yuchen He and Mike Heaton and Johannes Heidecke and Chris Hesse and Alan Hickey and Wade Hickey and Peter Hoeschele and Brandon Houghton and Kenny Hsu and Shengli Hu and Xin Hu and Joost Huizinga and Shantanu Jain and Shawn Jain and Joanne Jang and Angela Jiang and Roger Jiang and Haozhun Jin and Denny Jin and Shino Jomoto and Billie Jonn and Heewoo Jun and Tomer Kaftan and Łukasz Kaiser and Ali Kamali and Ingmar Kanitscheider and Nitish Shirish Keskar and Tabarak Khan and Logan Kilpatrick and Jong Wook Kim and Christina Kim and Yongjik Kim and Jan Hendrik Kirchner and Jamie Kiros and Matt Knight and Daniel Kokotajlo and Łukasz Kondraciuk and Andrew Kondrich and Aris Konstantinidis and Kyle Kosic and Gretchen Krueger and Vishal Kuo and Michael Lampe and Ikai Lan and Teddy Lee and Jan Leike and Jade Leung and Daniel Levy and Chak Ming Li and Rachel Lim and Molly Lin and Stephanie Lin and Mateusz Litwin and Theresa Lopez and Ryan Lowe and Patricia Lue and Anna Makanju and Kim Malfacini and Sam Manning and Todor Markov and Yaniv Markovski and Bianca Martin and Katie Mayer and Andrew Mayne and Bob McGrew and Scott Mayer McKinney and Christine McLeavey and Paul McMillan and Jake McNeil and David Medina and Aalok Mehta and Jacob Menick and Luke Metz and Andrey Mishchenko and Pamela Mishkin and Vinnie Monaco and Evan Morikawa and Daniel Mossing and Tong Mu and Mira Murati and Oleg Murk and David Mély and Ashvin Nair and Reiichiro Nakano and Rajeev Nayak and Arvind Neelakantan and Richard Ngo and Hyeonwoo Noh and Long Ouyang and Cullen O'Keefe and Jakub Pachocki and Alex Paino and Joe Palermo and Ashley Pantuliano and Giambattista Parascandolo and Joel Parish and Emy Parparita and Alex Passos and Mikhail Pavlov and Andrew Peng and Adam Perelman and Filipe de Avila Belbute Peres and Michael Petrov and Henrique Ponde de Oliveira Pinto and Michael and Pokorny and Michelle Pokrass and Vitchyr H. Pong and Tolly Powell and Alethea Power and Boris Power and Elizabeth Proehl and Raul Puri and Alec Radford and Jack Rae and Aditya Ramesh and Cameron Raymond and Francis Real and Kendra Rimbach and Carl Ross and Bob Rotsted and Henri Roussez and Nick Ryder and Mario Saltarelli and Ted Sanders and Shibani Santurkar and Girish Sastry and Heather Schmidt and David Schnurr and John Schulman and Daniel Selsam and Kyla Sheppard and Toki Sherbakov and Jessica Shieh and Sarah Shoker and Pranav Shyam and Szymon Sidor and Eric Sigler and Maddie Simens and Jordan Sitkin and Katarina Slama and Ian Sohl and Benjamin Sokolowsky and Yang Song and Natalie Staudacher and Felipe Petroski Such and Natalie Summers and Ilya Sutskever and Jie Tang and Nikolas Tezak and Madeleine B. Thompson and Phil Tillet and Amin Tootoonchian and Elizabeth Tseng and Preston Tuggle and Nick Turley and Jerry Tworek and Juan Felipe Cerón Uribe and Andrea Vallone and Arun Vijayvergiya and Chelsea Voss and Carroll Wainwright and Justin Jay Wang and Alvin Wang and Ben Wang and Jonathan Ward and Jason Wei and CJ Weinmann and Akila Welihinda and Peter Welinder and Jiayi Weng and Lilian Weng and Matt Wiethoff and Dave Willner and Clemens Winter and Samuel Wolrich and Hannah Wong and Lauren Workman and Sherwin Wu and Jeff Wu and Michael Wu and Kai Xiao and Tao Xu and Sarah Yoo and Kevin Yu and Qiming Yuan and Wojciech Zaremba and Rowan Zellers and Chong Zhang and Marvin Zhang and Shengjia Zhao and Tianhao Zheng and Juntang Zhuang and William Zhuk and Barret Zoph},
	year         = 2024,
	eprint       = {2303.08774},
	archiveprefix = {arXiv},
	primaryclass = {cs.CL}
}

@article{anthropic2024claude,
	title        = {Introducing {Claude} 3.5 {Sonnet}},
	author       = {{Anthropic}},
	year         = 2024,
	month        = {June},
	day          = 20,
}

@misc{deepseekai2025deepseekr1incentivizingreasoningcapability,
	title        = {DeepSeek-R1: Incentivizing Reasoning Capability in LLMs via Reinforcement Learning},
	author       = {DeepSeek-AI and Daya Guo and Dejian Yang and Haowei Zhang and Junxiao Song and Ruoyu Zhang and Runxin Xu and Qihao Zhu and Shirong Ma and Peiyi Wang and Xiao Bi and Xiaokang Zhang and Xingkai Yu and Yu Wu and Z. F. Wu and Zhibin Gou and Zhihong Shao and Zhuoshu Li and Ziyi Gao and Aixin Liu and Bing Xue and Bingxuan Wang and Bochao Wu and Bei Feng and Chengda Lu and Chenggang Zhao and Chengqi Deng and Chenyu Zhang and Chong Ruan and Damai Dai and Deli Chen and Dongjie Ji and Erhang Li and Fangyun Lin and Fucong Dai and Fuli Luo and Guangbo Hao and Guanting Chen and Guowei Li and H. Zhang and Han Bao and Hanwei Xu and Haocheng Wang and Honghui Ding and Huajian Xin and Huazuo Gao and Hui Qu and Hui Li and Jianzhong Guo and Jiashi Li and Jiawei Wang and Jingchang Chen and Jingyang Yuan and Junjie Qiu and Junlong Li and J. L. Cai and Jiaqi Ni and Jian Liang and Jin Chen and Kai Dong and Kai Hu and Kaige Gao and Kang Guan and Kexin Huang and Kuai Yu and Lean Wang and Lecong Zhang and Liang Zhao and Litong Wang and Liyue Zhang and Lei Xu and Leyi Xia and Mingchuan Zhang and Minghua Zhang and Minghui Tang and Meng Li and Miaojun Wang and Mingming Li and Ning Tian and Panpan Huang and Peng Zhang and Qiancheng Wang and Qinyu Chen and Qiushi Du and Ruiqi Ge and Ruisong Zhang and Ruizhe Pan and Runji Wang and R. J. Chen and R. L. Jin and Ruyi Chen and Shanghao Lu and Shangyan Zhou and Shanhuang Chen and Shengfeng Ye and Shiyu Wang and Shuiping Yu and Shunfeng Zhou and Shuting Pan and S. S. Li and Shuang Zhou and Shaoqing Wu and Shengfeng Ye and Tao Yun and Tian Pei and Tianyu Sun and T. Wang and Wangding Zeng and Wanjia Zhao and Wen Liu and Wenfeng Liang and Wenjun Gao and Wenqin Yu and Wentao Zhang and W. L. Xiao and Wei An and Xiaodong Liu and Xiaohan Wang and Xiaokang Chen and Xiaotao Nie and Xin Cheng and Xin Liu and Xin Xie and Xingchao Liu and Xinyu Yang and Xinyuan Li and Xuecheng Su and Xuheng Lin and X. Q. Li and Xiangyue Jin and Xiaojin Shen and Xiaosha Chen and Xiaowen Sun and Xiaoxiang Wang and Xinnan Song and Xinyi Zhou and Xianzu Wang and Xinxia Shan and Y. K. Li and Y. Q. Wang and Y. X. Wei and Yang Zhang and Yanhong Xu and Yao Li and Yao Zhao and Yaofeng Sun and Yaohui Wang and Yi Yu and Yichao Zhang and Yifan Shi and Yiliang Xiong and Ying He and Yishi Piao and Yisong Wang and Yixuan Tan and Yiyang Ma and Yiyuan Liu and Yongqiang Guo and Yuan Ou and Yuduan Wang and Yue Gong and Yuheng Zou and Yujia He and Yunfan Xiong and Yuxiang Luo and Yuxiang You and Yuxuan Liu and Yuyang Zhou and Y. X. Zhu and Yanhong Xu and Yanping Huang and Yaohui Li and Yi Zheng and Yuchen Zhu and Yunxian Ma and Ying Tang and Yukun Zha and Yuting Yan and Z. Z. Ren and Zehui Ren and Zhangli Sha and Zhe Fu and Zhean Xu and Zhenda Xie and Zhengyan Zhang and Zhewen Hao and Zhicheng Ma and Zhigang Yan and Zhiyu Wu and Zihui Gu and Zijia Zhu and Zijun Liu and Zilin Li and Ziwei Xie and Ziyang Song and Zizheng Pan and Zhen Huang and Zhipeng Xu and Zhongyu Zhang and Zhen Zhang},
	year         = 2025,
	eprint       = {2501.12948},
	archiveprefix = {arXiv},
}

@inproceedings{ferrando2025do,
	title        = {Do I Know This Entity? Knowledge Awareness and Hallucinations in Language Models},
	author       = {Javier Ferrando and Oscar Balcells Obeso and Senthooran Rajamanoharan and Neel Nanda},
	year         = 2025,
	booktitle    = {ICLR},
}

@inproceedings{demircan2025sparse,
	title        = {Sparse Autoencoders Reveal Temporal Difference Learning in Large Language Models},
	author       = {Can Demircan and Tankred Saanum and Akshay Kumar Jagadish and Marcel Binz and Eric Schulz},
	year         = 2025,
	booktitle    = {ICLR},
}

@inproceedings{second,
	title        = {SECOND: Mitigating Perceptual Hallucination in Vision-Language Models via Selective and Contrastive Decoding},
	author       = {Park, Woohyeon and Kim, Woojin and Kim, Jaeik and Do, Jaeyoung},
	year         = 2025,
	booktitle    = {ICML}
}

@inproceedings{lou2025sae,
	title        = {SAE-V: Interpreting Multimodal Models for Enhanced Alignment},
	author       = {Lou, Hantao and Li, Changye and Ji, Jiaming and Yang, Yaodong},
	year         = 2025,
	booktitle    = {ICML}
}

@article{zhang2024large,
	title        = {Large multi-modal models can interpret features in large multi-modal models},
	author       = {Zhang, Kaichen and Shen, Yifei and Li, Bo and Liu, Ziwei},
	year         = 2024,
	journal      = {arXiv preprint arXiv:2411.14982}
}

@article{wu2025interpreting,
	title        = {Interpreting and steering llms with mutual information-based explanations on sparse autoencoders},
	author       = {Wu, Xuansheng and Yuan, Jiayi and Yao, Wenlin and Zhai, Xiaoming and Liu, Ninghao},
	year         = 2025,
	journal      = {arXiv preprint arXiv:2502.15576}
}

@article{gallifant2025sparse,
	title        = {Sparse autoencoder features for classifications and transferability},
	author       = {Gallifant, Jack and Chen, Shan and Sasse, Kuleen and Aerts, Hugo and Hartvigsen, Thomas and Bitterman, Danielle S},
	year         = 2025,
	journal      = {arXiv preprint arXiv:2502.11367}
}

@inproceedings{nanda2024progress,
	title        = {Progress update\# 1 from the gdm mech interp team: Full update},
	author       = {Nanda, Neel and Conmy, Arthur and Smith, Lewis and Rajamanoharan, Senthooran and Lieberum, Tom and Kram{\'a}r, J{\'a}nos and Varma, Vikrant},
	year         = 2024,
	booktitle    = {AI Alignment Forum}
}

@inproceedings{liu2024llava1.5,
	title        = {Improved baselines with visual instruction tuning},
	author       = {Liu, Haotian and Li, Chunyuan and Li, Yuheng and Lee, Yong Jae},
	year         = 2024,
	booktitle    = {CVPR},
	pages        = {26296--26306}
}

@article{pai,
	title        = {Paying more attention to image: A training-free method for alleviating hallucination in lvlms},
	author       = {Liu, Shi and Zheng, Kecheng and Chen, Wei},
	year         = 2024,
	journal      = {arXiv preprint arXiv:2407.21771}
}

@article{comanici2025gemini,
	title        = {Gemini 2.5: Pushing the frontier with advanced reasoning, multimodality, long context, and next generation agentic capabilities},
	author       = {Comanici, Gheorghe and Bieber, Eric and Schaekermann, Mike and Pasupat, Ice and Sachdeva, Noveen and Dhillon, Inderjit and Blistein, Marcel and Ram, Ori and Zhang, Dan and Rosen, Evan and others},
	year         = 2025,
	journal      = {arXiv preprint arXiv:2507.06261}
}

@article{yang2025qwen3,
	title        = {Qwen3 technical report},
	author       = {Yang, An and Li, Anfeng and Yang, Baosong and Zhang, Beichen and Hui, Binyuan and Zheng, Bo and Yu, Bowen and Gao, Chang and Huang, Chengen and Lv, Chenxu and others},
	year         = 2025,
	journal      = {arXiv preprint arXiv:2505.09388}
}

@article{zhai2024halleControl,
	title        = {HallE-Control: Controlling Object Hallucination in Large Multimodal Models},
	author       = {Bohan Zhai and Shijia Yang and Chenfeng Xu and Sheng Shen and Kurt Keutzer and Chunyuan Li and Manling Li},
	year         = 2024,
	journal      = {arXiv preprint arXiv:2310.01779},
}

@article{icd,
	title        = {Mitigating hallucinations in large vision-language models with instruction contrastive decoding},
	author       = {Wang, Xintong and Pan, Jingheng and Ding, Liang and Biemann, Chris},
	year         = 2024,
	journal      = {arXiv preprint arXiv:2403.18715}
}

@article{sae_survey,
	title        = {A survey on sparse autoencoders: Interpreting the internal mechanisms of large language models},
	author       = {Shu, Dong and Wu, Xuansheng and Zhao, Haiyan and Rai, Daking and Yao, Ziyu and Liu, Ninghao and Du, Mengnan},
	year         = 2025,
	journal      = {arXiv preprint arXiv:2503.05613}
}

@inproceedings{yu2024hallucidoctor,
	title        = {Hallucidoctor: Mitigating hallucinatory toxicity in visual instruction data},
	author       = {Yu, Qifan and Li, Juncheng and Wei, Longhui and Pang, Liang and Ye, Wentao and Qin, Bosheng and Tang, Siliang and Tian, Qi and Zhuang, Yueting},
	year         = 2024,
	booktitle    = {CVPR},
	pages        = {12944--12953}
}

@inproceedings{llava,
  title={Visual instruction tuning},
  author={Liu, Haotian and Li, Chunyuan and Wu, Qingyang and Lee, Yong Jae},
  booktitle={NeurIPS},
  year={2023}
}

@misc{llavanext,
	title        = {Llavanext: Improved reasoning, ocr, and world knowledge},
	author       = {Liu, Haotian and Li, Chunyuan and Li, Yuheng and Li, Bo and Zhang, Yuanhan and Shen, Sheng and Lee, Yong Jae},
	year         = 2024
}

@article{amber,
	title        = {Amber: An llm-free multi-dimensional benchmark for mllms hallucination evaluation},
	author       = {Wang, Junyang and Wang, Yuhang and Xu, Guohai and Zhang, Jing and Gu, Yukai and Jia, Haitao and Wang, Jiaqi and Xu, Haiyang and Yan, Ming and Zhang, Ji and others},
	year         = 2023,
	journal      = {arXiv preprint arXiv:2311.07397}
}

@inproceedings{hallusionbench,
	title        = {Hallusionbench: an advanced diagnostic suite for entangled language hallucination and visual illusion in large vision-language models},
	author       = {Guan, Tianrui and Liu, Fuxiao and Wu, Xiyang and Xian, Ruiqi and Li, Zongxia and Liu, Xiaoyu and Wang, Xijun and Chen, Lichang and Huang, Furong and Yacoob, Yaser and others},
	year         = 2024,
	booktitle    = {CVPR},
	pages        = {14375--14385}
}

@article{kang2025see_attention_sinks,
	title        = {See what you are told: Visual attention sink in large multimodal models},
	author       = {Kang, Seil and Kim, Jinyeong and Kim, Junhyeok and Hwang, Seong Jae},
	year         = 2025,
	journal      = {arXiv:2503.03321}
}

@article{sun2024massive_activations_attention_sinks,
	title        = {Massive activations in large language models},
	author       = {Sun, Mingjie and Chen, Xinlei and Kolter, J Zico and Liu, Zhuang},
	year         = 2024,
	journal      = {arXiv:2402.17762}
}

@inproceedings{kaduri2025s_cvpr25,
	title        = {What's in the Image? A Deep-Dive into the Vision of Vision Language Models},
	author       = {Kaduri, Omri and Bagon, Shai and Dekel, Tali},
	year         = 2025,
	booktitle    = {CVPR},
	pages        = {14549--14558}
}

@string(CVPR= {IEEE Conf. Comput. Vis. Pattern Recog.})

@string(NIPS= {Adv. Neural Inform. Process. Syst.})

@string(ICLR = {Int. Conf. Learn. Represent.})

@string(AAAI = {AAAI})

@string(CVPR  = {CVPR})

@string(NIPS  = {NeurIPS})

@string(ICLR  = {ICLR})

@article{lyu2025revealing_guangtao_lmm,
	title        = {Revealing Perception and Generation Dynamics in LVLMs: Mitigating Hallucinations via Validated Dominance Correction},
	author       = {Lyu, Guangtao and Cheng, Xinyi and Xu, Chenghao and Liu, Qi and Yang, Muli and Fang, Fen and Chen, Huilin and Yan, Jiexi and Yang, Xu and Deng, Cheng},
	year         = 2025,
	journal      = {arXiv:2512.18813}
}

@inproceedings{agla,
	title        = {Mitigating object hallucinations in large vision-language models with assembly of global and local attention},
	author       = {An, Wenbin and Tian, Feng and Leng, Sicong and Nie, Jiahao and Lin, Haonan and Wang, QianYing and Chen, Ping and Zhang, Xiaoqin and Lu, Shijian},
	year         = 2025,
	booktitle    = {CVPR}
}

@inproceedings{hallu_attention_lens,
	title        = {Devils in middle layers of large vision-language models: Interpreting, detecting and mitigating object hallucinations via attention lens},
	author       = {Jiang, Zhangqi and Chen, Junkai and Zhu, Beier and Luo, Tingjin and Shen, Yankun and Yang, Xu},
	year         = 2025,
	booktitle    = {CVPR}
}

@inproceedings{dola_looking_twice_memvr_ffn_iternal,
	title        = {Look Twice Before You Answer: Memory-Space Visual Retracing for Hallucination Mitigation in Multimodal Large Language Models},
	author       = {Zou, Xin and Wang, Yizhou and Yan, Yibo and Huang, Sirui and Zheng, Kening and Chen, Junkai and Tang, Chang and Hu, Xuming},
	year         = 2025,
	booktitle    = {ICML}
}

@article{Seeing_but_Not_Believing,
	title        = {Seeing but Not Believing: Probing the Disconnect Between Visual Attention and Answer Correctness in VLMs},
	author       = {Liu, Zhining and Chen, Ziyi and Liu, Hui and Luo, Chen and Tang, Xianfeng and Wang, Suhang and Zeng, Joy and Dai, Zhenwei and Shi, Zhan and Wei, Tianxin and others},
	year         = 2025,
	journal      = {arXiv:2510.17771}
}

@inproceedings{yin2025clearsight_vaf,
	title        = {ClearSight: Visual Signal Enhancement for Object Hallucination Mitigation in Multimodal Large Language Models},
	author       = {Yin, Hao and Si, Guangzong and Wang, Zilei},
	year         = 2025,
	booktitle    = {CVPR},
	pages        = {14625--14634}
}

@article{xiao2023efficient_attention_sinks_ori,
	title        = {Efficient streaming language models with attention sinks},
	author       = {Xiao, Guangxuan and Tian, Yuandong and Chen, Beidi and Han, Song and Lewis, Mike},
	year         = 2023,
	journal      = {arXiv:2309.17453}
}

@inproceedings{register_need,
	title        = {Vision Transformers Need Registers},
	author       = {Darcet, Timoth{\'e}e and Oquab, Maxime and Mairal, Julien and Bojanowski, Piotr},
	year         = 2024,
	booktitle    = {ICLR}
}

@inproceedings{gurari2018vizwiz,
	title        = {Vizwiz grand challenge: Answering visual questions from blind people},
	author       = {Gurari, Danna and Li, Qing and Stangl, Abigale J and Guo, Anhong and Lin, Chi and Grauman, Kristen and Luo, Jiebo and Bigham, Jeffrey P},
	year         = 2018,
	booktitle    = {CVPR},
	pages        = {3608--3617}
}

@article{yu2023mmvet,
	title        = {Mm-vet: Evaluating large multimodal models for integrated capabilities},
	author       = {Yu, Weihao and Yang, Zhengyuan and Li, Linjie and Wang, Jianfeng and Lin, Kevin and Liu, Zicheng and Wang, Xinchao and Wang, Lijuan},
	year         = 2023,
	journal      = {arXiv:2308.02490}
}

@inproceedings{register_dont_need,
	title        = {Vision Transformers Don't Need Trained Registers},
	author       = {Jiang, Nick and Dravid, Amil and Efros, Alexei and Gandelsman, Yossi},
	year         = 2025,
	booktitle    = {NeurIPS}
}
\bibliographystyle{icml2026}

%%%%%%%%%%%%%%%%%%%%%%%%%%%%%%%%%%%%%%%%%%%%%%%%%%%%%%%%%%%%%%%%%%%%%%%%%%%%%%%
%%%%%%%%%%%%%%%%%%%%%%%%%%%%%%%%%%%%%%%%%%%%%%%%%%%%%%%%%%%%%%%%%%%%%%%%%%%%%%%
% APPENDIX
%%%%%%%%%%%%%%%%%%%%%%%%%%%%%%%%%%%%%%%%%%%%%%%%%%%%%%%%%%%%%%%%%%%%%%%%%%%%%%%
%%%%%%%%%%%%%%%%%%%%%%%%%%%%%%%%%%%%%%%%%%%%%%%%%%%%%%%%%%%%%%%%%%%%%%%%%%%%%%%
\newpage
\appendix
\onecolumn

\begin{figure*}[t]
    \centering
    \includegraphics[width=0.95\linewidth]{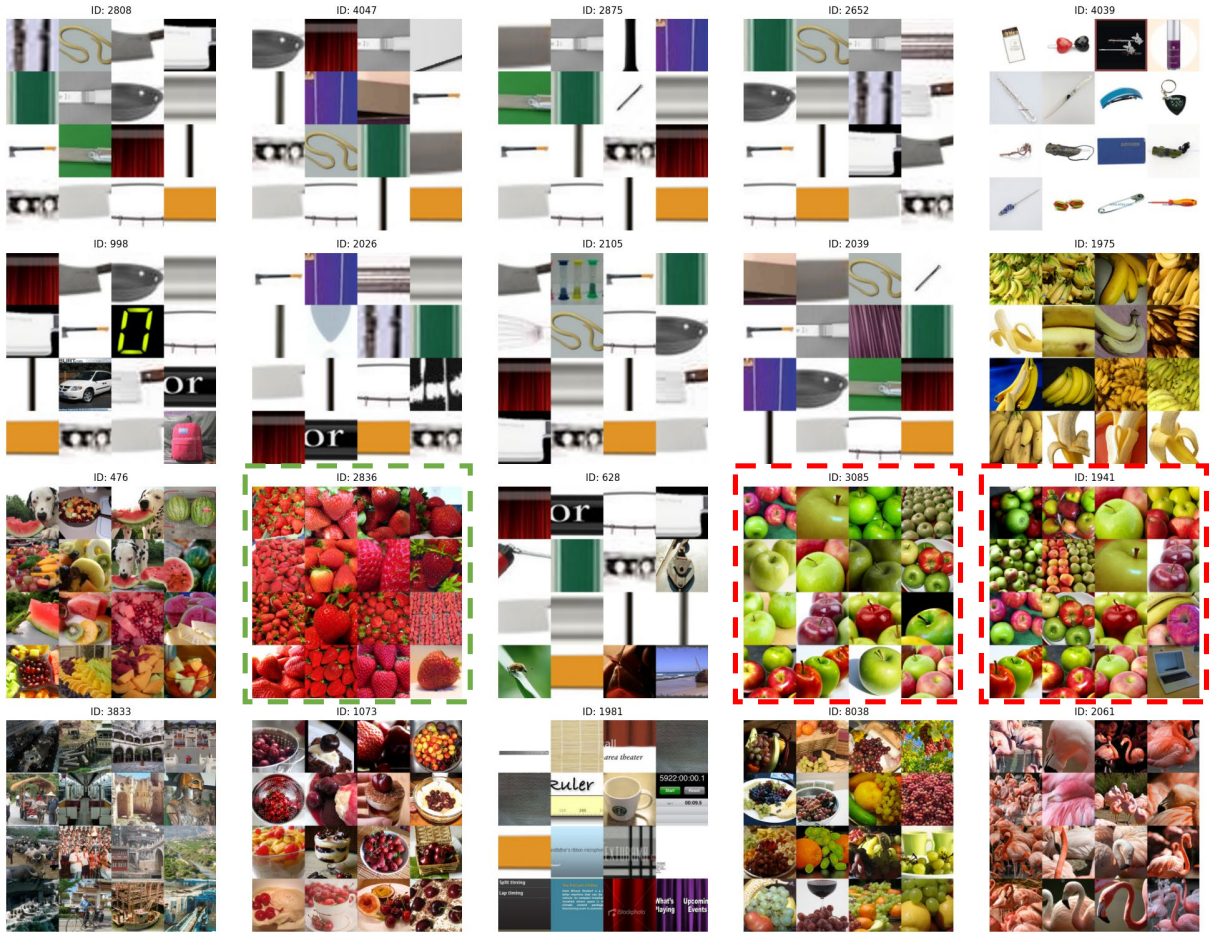}
\caption{
\textbf{Attribute hallucination caused by competing concept activations.}
The image shows a fruit bowl containing multiple fruits, including a black apple.
When asked “What is the color of the apple?”, the model incorrectly answers “red,” exhibiting an attribute-level hallucination.
Neuron-level analysis reveals that a strawberry-related neuron (ID 2836; \textcolor{green}{green boxes}) becomes abnormally activated and dominates the representation, overwhelming evidence from apple-related neurons (\textcolor{red}{red boxes}).
By suppressing neuron 2836 or enhancing apple-specific neurons (IDs 3085 and 1941), the model correctly outputs “black.”
This example illustrates how hallucinations can arise from spurious dominance of irrelevant semantic neurons and how targeted neuron modulation restores correct visual grounding.
}

    \label{fig:app_case_black_apple}
\end{figure*}

\begin{figure*}[t]
    \centering
    \includegraphics[width=0.9\linewidth]{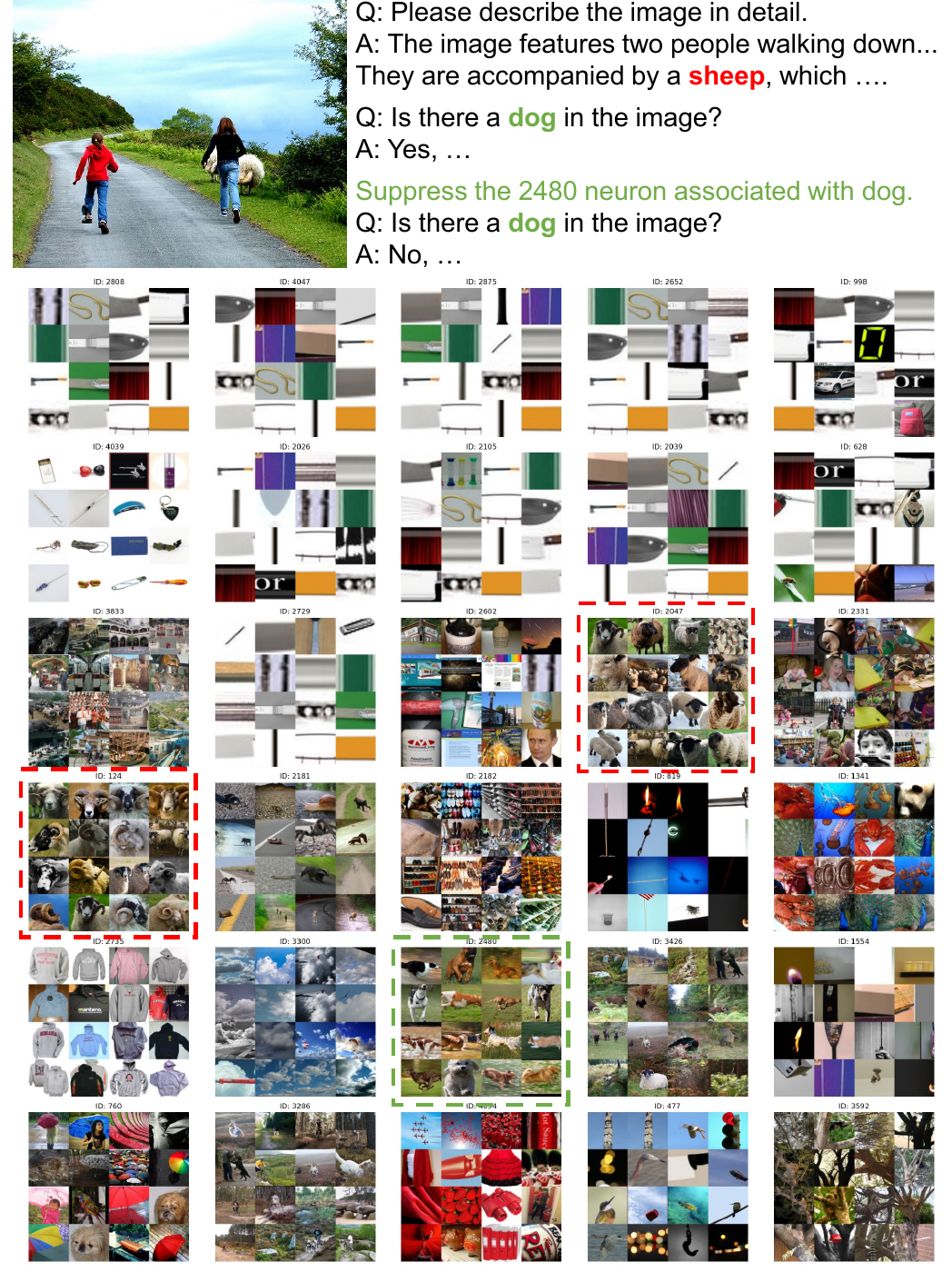}
\caption{
\textbf{Object hallucination induced by spurious activation under occlusion.}
The activation maps indicate that the model primarily focuses on the sheep, with sheep-related neurons strongly activated (\textcolor{red}{red boxes}), allowing it to describe the image correctly. However, because the sheep is partially occluded, neurons linked to other animal concepts, such as the dog (\textcolor{green}{green boxes}), are also activated. While these additional activations do not affect general image description, they become problematic when the model is asked a concept-specific question such as “Is there a dog in the image?” leading it to incorrectly answer “yes.” After suppressing the dog-related neuron 2480 by setting its weight to –10, the model correctly responds “no.” This example highlights that extra activation of irrelevant concepts can mislead the model under targeted queries and that neuron-level modulation provides an effective means to suppress such spurious signals and mitigate hallucinations.
}

    \label{fig:app_case_sheep}
\end{figure*}

\begin{figure*}[t]
    \centering
    \includegraphics[width=0.9\linewidth]{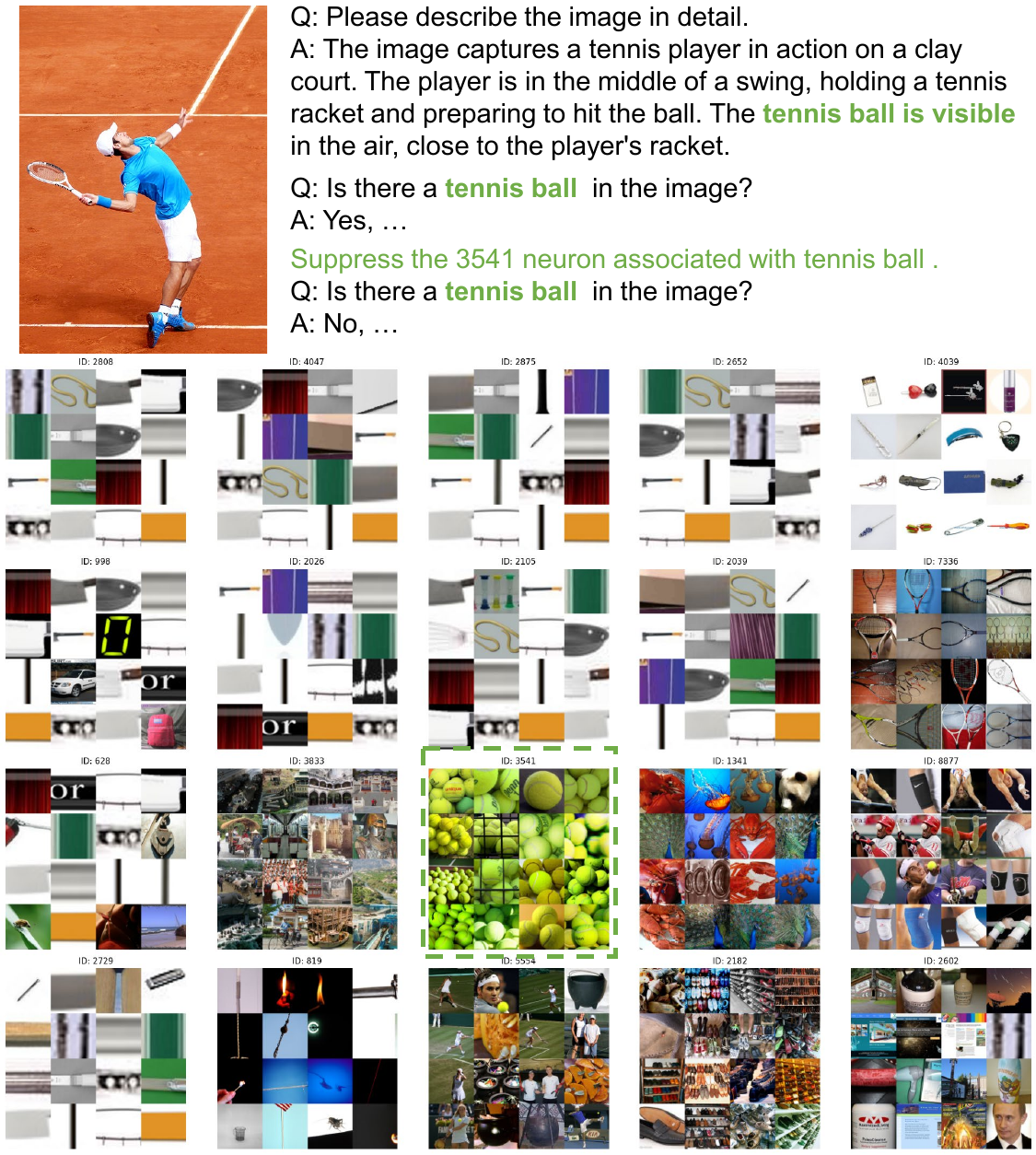}
\caption{
\textbf{Object hallucination caused by spurious concept injection.}
The image depicts a tennis player swinging a racket on a clay court, but \emph{no tennis ball is present}.
Despite the absence of visual evidence, the model hallucinates a tennis ball in its description, claiming that “the tennis ball is visible in the air near the racket.”
Neuron-level analysis attributes this error to abnormal activation of a tennis-ball–specific neuron (ID 3541), which injects an unsupported object concept into the visual representation.
After suppressing this neuron, the model no longer hallucinates the tennis ball and correctly answers “no” when asked about its presence.
This example highlights how isolated spurious neuron activations can directly lead to object-level hallucinations.
}

    \label{fig:app_case_tennis_ball}
\end{figure*}

\begin{figure*}[t]
    \centering
    \includegraphics[width=0.95\linewidth]{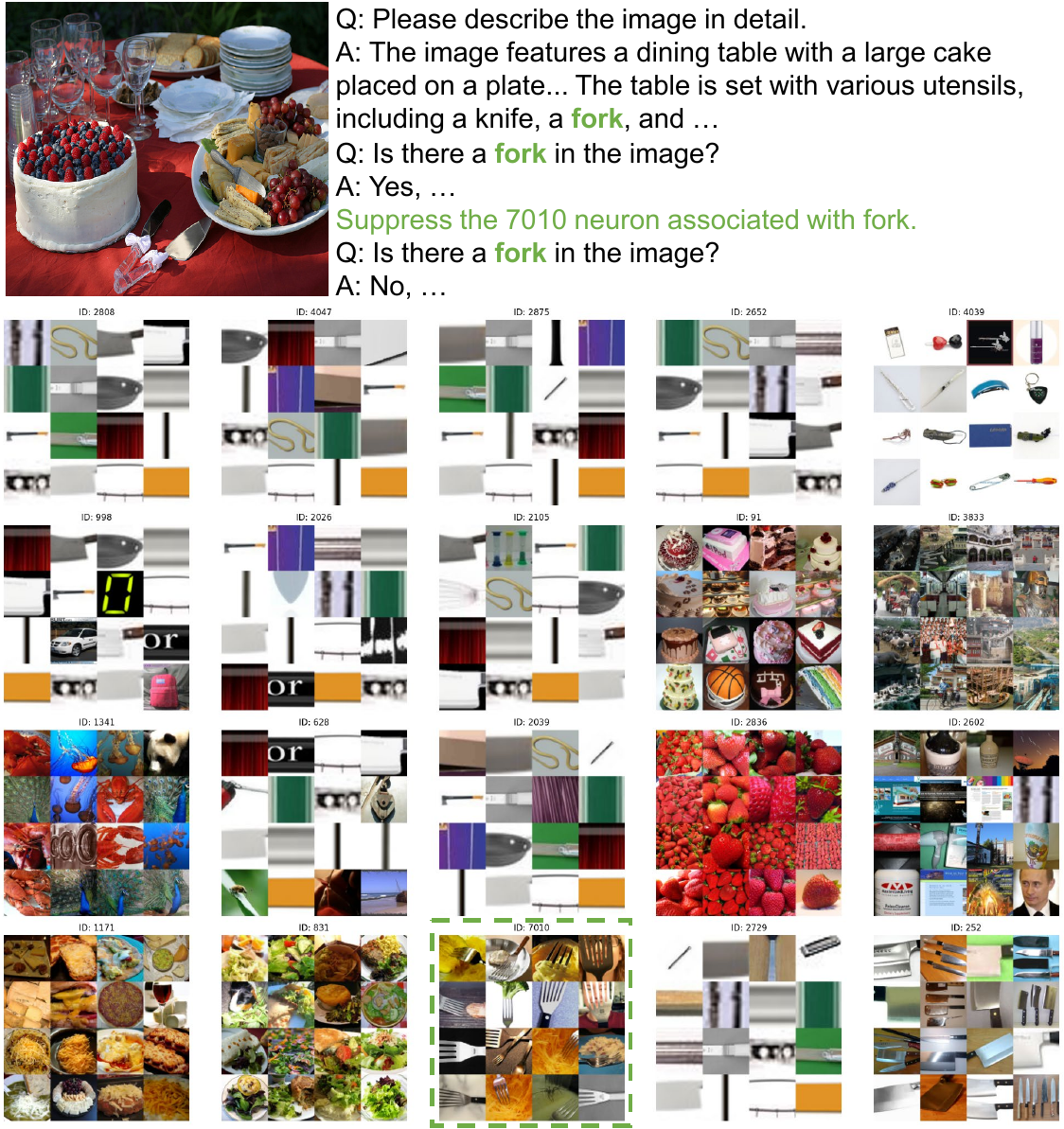}
\caption{
\textbf{Object hallucination driven by overactive utensil-related neurons.}
The image depicts a dining table filled with food and drinks, including a cake, fruit, cookies, and crackers, but \emph{no fork is present in the scene}.
Nevertheless, the model hallucinates a fork in its image description and subsequently answers “yes” when asked whether a fork appears in the image.
Neuron-level inspection reveals that this error is caused by abnormal activation of a fork-related concept neuron (ID 7010), which introduces an unsupported utensil concept into the visual representation.
After suppressing this neuron, the model no longer mentions a fork and correctly answers “no” to the concept-specific query.
This example further illustrates that object hallucinations in LVLMs can originate from isolated yet overactive concept neurons, and that targeted neuron-level modulation effectively restores correct visual grounding and factual consistency.
}

    \label{fig:app_case_fork}
\end{figure*}

\section{Detailed Neuron-Level Hallucination Analysis and Mitigation}
\label{sec:detailed_hallucinations}

\Cref{fig:app_case_black_apple,fig:app_case_sheep,fig:app_case_tennis_ball,fig:app_case_fork} present representative examples illustrating how hallucinations in LVLMs arise from abnormal, competing, or context-driven neuron activations under different captioning and targeted questioning scenarios. These cases further demonstrate that targeted neuron-level interventions can effectively suppress spurious signals and improve factual reliability and visual grounding in a controllable and interpretable manner.

In the black-apple example (\Cref{fig:app_case_black_apple}), the model is asked a factual attribute question: “What is the color of the apple?” Although the image clearly shows a black apple, the model initially predicts “red.” Neuron-level analysis reveals that neuron 2836, associated with the concept of red strawberries, exhibits abnormally strong activation. This spurious activation dominates apple-related neurons, leading the model to prioritize an irrelevant concept and produce a hallucinated prediction. By selectively suppressing neuron 2836 or amplifying apple-specific neurons (IDs 3085 and 1941), the activation distribution shifts toward the correct visual evidence, and the model correctly outputs “black.” This example illustrates how attribute-level hallucinations can emerge from misaligned concept activations and how targeted neuron modulation can restore correct visual grounding while offering an interpretable explanation of the failure mode.

In the sheep-and-dog example (\Cref{fig:app_case_sheep}), the model produces a generally accurate image description by focusing on the sheep, with sheep-related neurons strongly activated. However, due to visual ambiguity introduced by partial occlusion, neurons associated with other animal concepts, such as the dog, are also activated. While such activations do not affect general image captioning, they become problematic when the model is asked a concept-specific question, e.g., “Is there a dog in the image?” In this case, the presence of dog-related neuron activation leads the model to incorrectly answer “yes,” despite the absence of a dog. By strongly suppressing the dog-related neuron (ID 2480), the model correctly answers “no.” This case demonstrates that even weak or secondary activations of irrelevant concepts can induce hallucinations under targeted queries, and that neuron-level interventions provide an effective mechanism for suppressing such misleading signals.

In the tennis-ball example (\Cref{fig:app_case_tennis_ball}), the model exhibits an object-level hallucination caused by spurious concept injection. The image depicts a tennis player swinging a racket on a clay court, yet no tennis ball is present. Despite the lack of visual evidence, the model hallucinates a tennis ball and claims that it is visible near the racket. Neuron-level inspection attributes this error to abnormal activation of a tennis-ball–specific neuron (ID 3541), which introduces an unsupported object concept into the visual representation. Unlike cases involving competing concepts, this hallucination is driven by a single dominant neuron whose activation is not grounded in the input image. After suppressing neuron 3541, the model no longer hallucinates the tennis ball and correctly answers “no” when queried about its presence. This example highlights how isolated spurious neuron activations can directly inject non-existent objects into the model’s output.

In the fork example (\Cref{fig:app_case_fork}), the model hallucinates an object due to overactivation of a contextually plausible but visually absent concept. The image shows a dining table containing various food items, such as cake, fruit, cookies, and crackers, but no fork is present. Nevertheless, the model includes a fork in its description and answers “yes” when asked whether a fork appears in the image. Neuron-level analysis reveals that this behavior is driven by abnormal activation of a fork-related neuron (ID 7010), likely influenced by strong co-occurrence priors between dining scenes and utensils. This neuron injects a utensil concept into the representation despite the absence of supporting visual evidence. Suppressing neuron 7010 removes the hallucinated fork and restores the correct response. This case illustrates how contextual priors encoded at the neuron level can lead to object hallucinations, and how targeted suppression can effectively recover factual consistency.

Across all examples, a consistent pattern is revealed: hallucinations in LVLMs frequently result from spurious, competing, or context-driven activations of concept-specific neurons that are insufficiently grounded in the visual input. Such activations may be triggered by visual ambiguity, strong semantic priors, or dominant but irrelevant concepts. By selectively amplifying neurons aligned with the true visual evidence and suppressing misleading ones, the internal representations become more consistent with the input image. This alignment reduces hallucinations, improves interpretability, and enhances controllability, providing a direct mechanism to mitigate errors at the neuron level.

Overall, these analyses highlight the dual value of neuron-level examination. First, it provides transparent insights into the internal mechanisms that give rise to hallucinations. Second, it enables actionable and fine-grained interventions that directly modulate model behavior. Collectively, these findings demonstrate that interpretable neuron-level interventions not only enhance factual consistency and visual grounding, but also serve as an effective strategy for mitigating hallucinations in LVLMs.

\clearpage

\section{Additional Neuron Visualizations}
\label{sec:app_neuron_vis_more}
\begin{figure*}[t]
    \centering
    \includegraphics[width=0.99\linewidth]{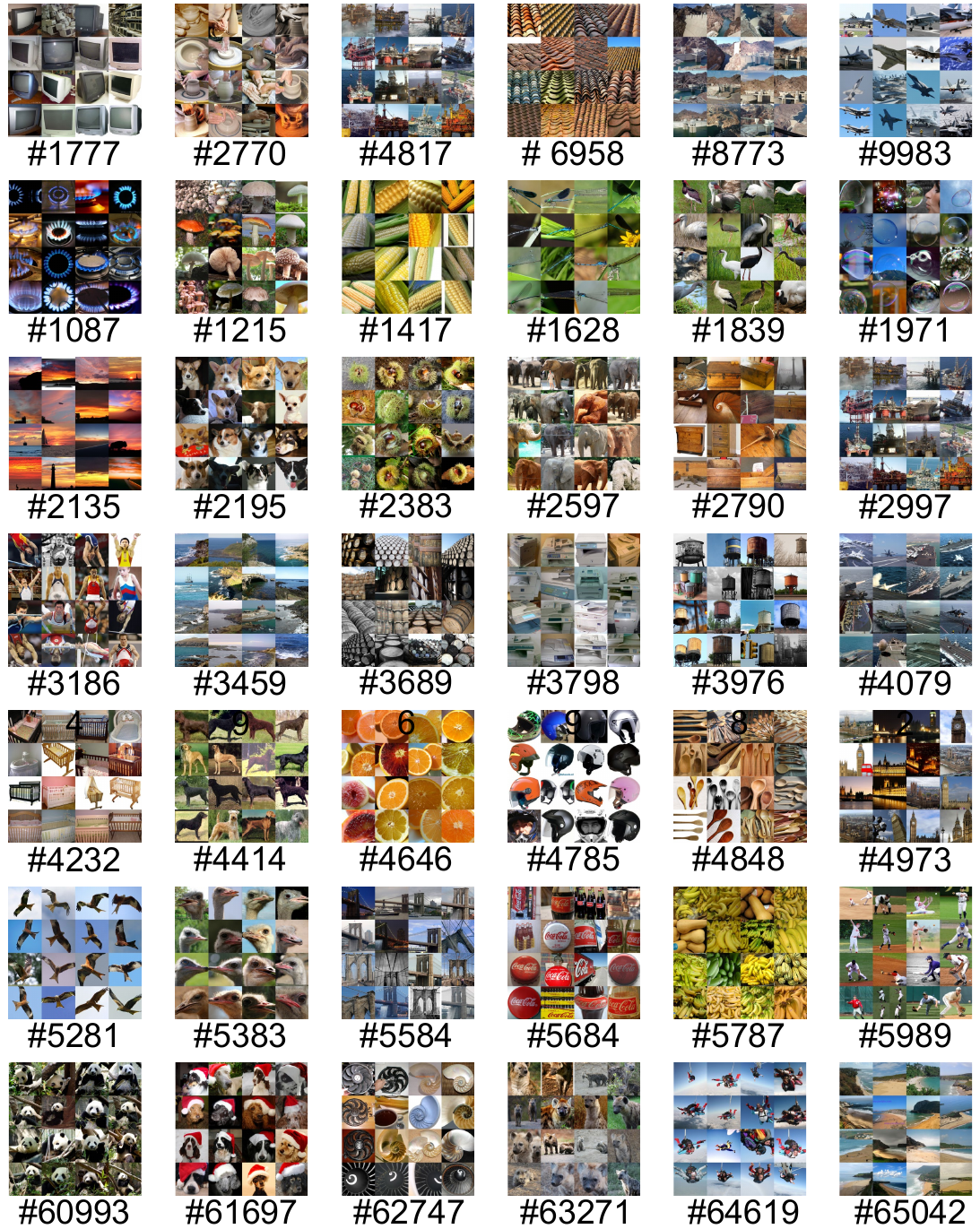}
    \vspace{-5pt}
    \caption{Additional visualizations of neurons learned by the SAE. The neurons exhibit diverse semantics, ranging from specific objects (e.g., corn \#14174, oranges \#46469, Christmas-hat dogs \#61697) to abstract structural patterns (e.g., spirals or fan-like shapes \#62747). This diversity demonstrates the interpretability of the internal representation space and provides a strong foundation for explaining and steering LVLMs through neuron-level interventions.}
\vspace{-5pt}
    \label{fig:more_neurons_vis}
\end{figure*}

\Cref{fig:more_neurons_vis} presents additional examples of neurons discovered by our sparse autoencoder (SAE). Many neurons exhibit strong associations with concrete objects or concepts, such as \#14174 for corn, \#46469 for oranges, and \#61697 for dogs wearing Christmas hats. Beyond object-level semantics, some neurons capture more abstract structural cues, such as \#62747, which consistently responds to spiral or fan-shaped patterns.  
These examples demonstrate the richness and diversity of the learned neuron space, ranging from fine-grained objects to higher-level structural abstractions. Such diversity not only enhances the interpretability of internal visual representations but also provides a strong foundation for precise neuron-level interventions, thereby facilitating both mechanistic understanding and controllable steering of LVLM outputs.

\section{Additional Neuron Analysis}
\label{sec:app_analyse_sae_neuron_vis}
To gain deeper insights into the functional roles of individual neurons in LVLM visual representations, we perform both image-level and patch-level analyses. These qualitative results complement the quantitative findings in \Cref{sec:analyse_sae}, providing a more intuitive understanding of how neurons encode semantic information.

\begin{figure*}[t]
    \centering
    \includegraphics[width=0.92\linewidth]{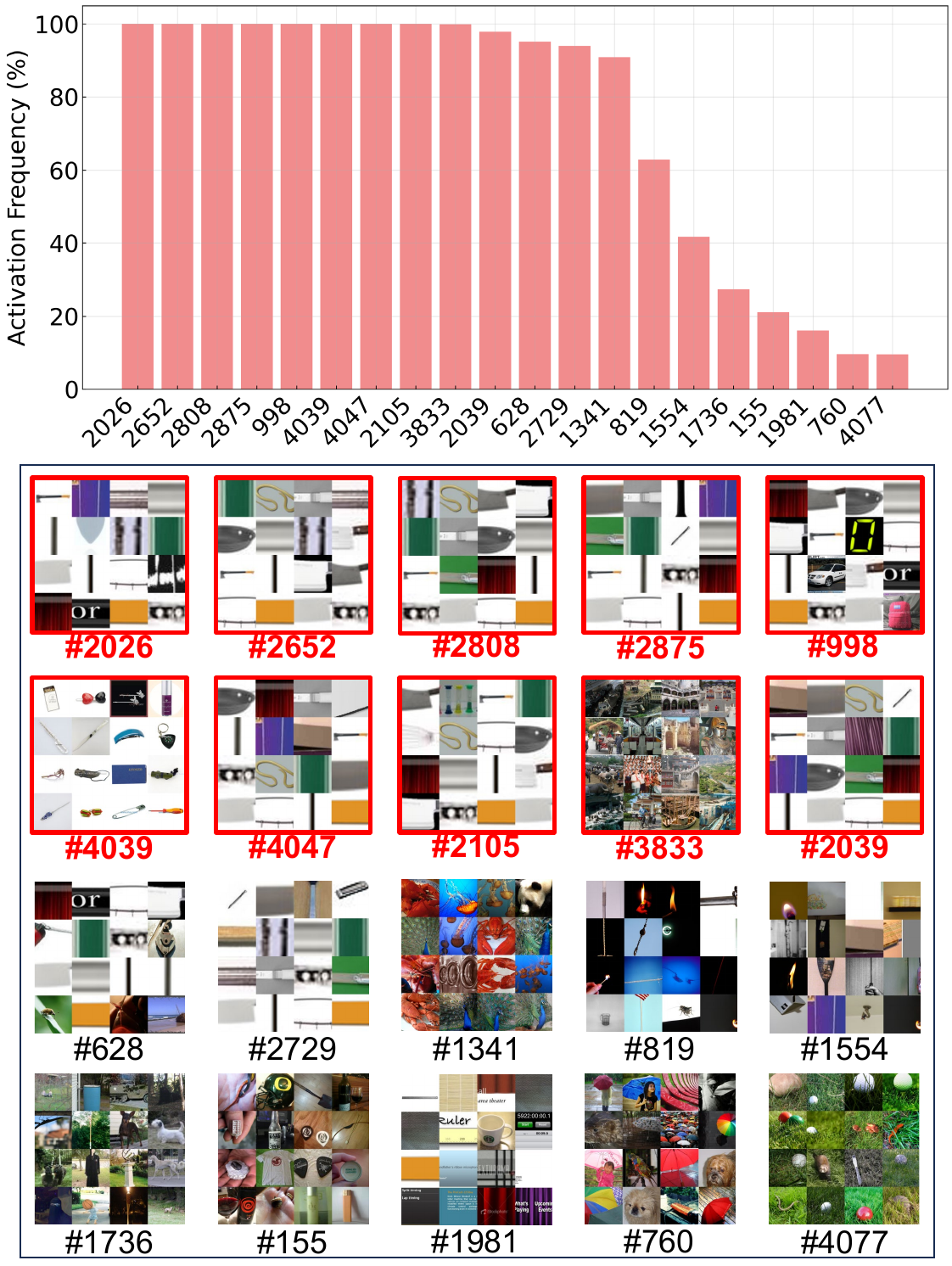}
\caption{Image-level neuron analysis and visualization.
The top panel shows neurons with consistently high activation rates across different images, while the bottom panel visualizes the top-activated images for each neuron. Neurons highlighted by \textcolor{red}{red boxes} are \emph{always-on neurons}, which tend to respond to recurring textures or small visual patterns and encode similar global information across images. These red-highlighted neurons are selected for suppression in CNS via ANS. Suppressing such always-on neurons helps emphasize image-specific objects and provides an interpretable basis for our neuron-level intervention.}    \vspace{-6pt}
    \label{fig:app_image_level_neurons_vis}
\end{figure*}

\paragraph{Image-Level Analysis and Visualizations.}
\Cref{fig:app_image_level_neurons_vis} highlights neurons with consistently high activation across different images. We observe that a small subset of ``always-on'' neurons remain persistently active regardless of image content, often encoding recurring textures or repetitive small objects rather than scene-specific information. The bottom panel further visualizes the top-activated images for each neuron, confirming that these neurons capture similar global patterns across diverse inputs. Within CNS, we reduce their disproportionate influence via Always-on Neuron Suppression (ANS), which decreases redundancy, emphasizes image-specific content, and improves the interpretability of downstream neuron-level interventions.

\begin{figure*}[t]
    \centering
    \includegraphics[width=0.95\linewidth]{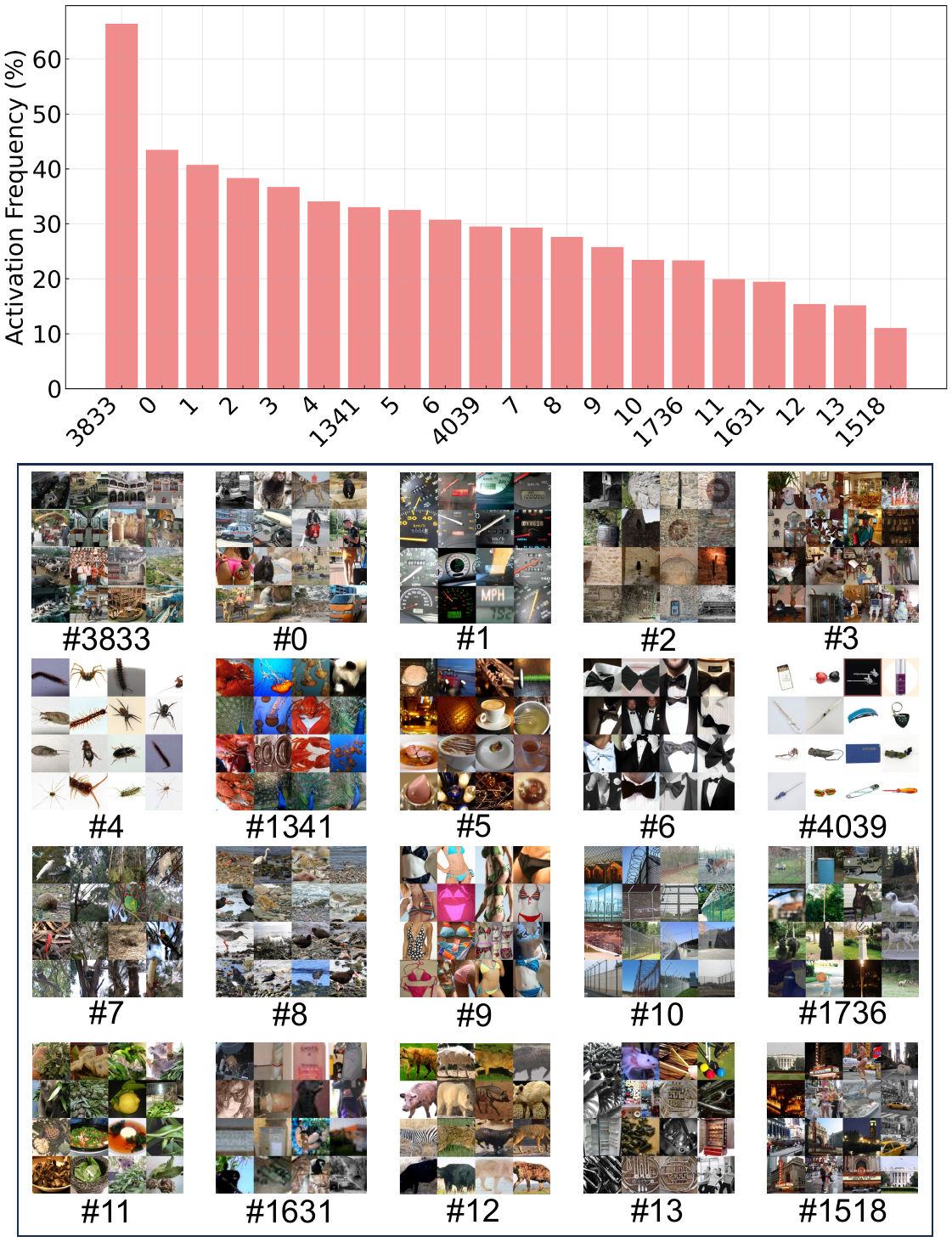}
    \caption{Patch-Level neurons analysis and visualization. At the patch level, neurons often capture concrete, localized concepts. Activation patterns show that most neurons reliably represent specific visual features, supporting fine-grained neuron-level interventions.}
    \vspace{-5pt}
    \label{fig:app_patch_neuron_vis}
\end{figure*}

\paragraph{Patch-Level Analysis and Visualizations.}
\Cref{fig:app_patch_neuron_vis} illustrates neuron activations at the patch level. Unlike always-on neurons, most patch-level neurons respond reliably to localized, semantically meaningful concepts, such as distinctive textures or object parts. This indicates that individual neurons often encode interpretable, fine-grained features, which are particularly well-suited for targeted interventions. By selectively modulating these neurons, we can directly influence which visual concepts are emphasized or suppressed in LVLM outputs, enabling fine-grained and interpretable control.

\paragraph{Summary.}
Together, the image-level and patch-level analyses reveal a dual organization of neuron activations: broadly active global features and selectively tuned local features. This dual perspective underpins our CNS approach, where targeted neuron-level interventions enable controllable and interpretable mitigation of hallucinations, while also deepening mechanistic insights into LVLM visual processing.

\section{Noise-Induced Disruption of Internal Visual Features Leading to Hallucinations}
\label{sec:app_neuron_change_vis}
In \Cref{sec:analyse_sae}, we quantitatively analyzed how noise perturbs internal visual features, causing neuron activations to shift and destabilize. These disruptions reshape the semantics of visual representations, inducing hallucinations and degrading LVLM performance (see \Cref{fig:explore_neuron_change}).

\begin{figure*}[t]
    \centering
    \includegraphics[width=0.95\linewidth]{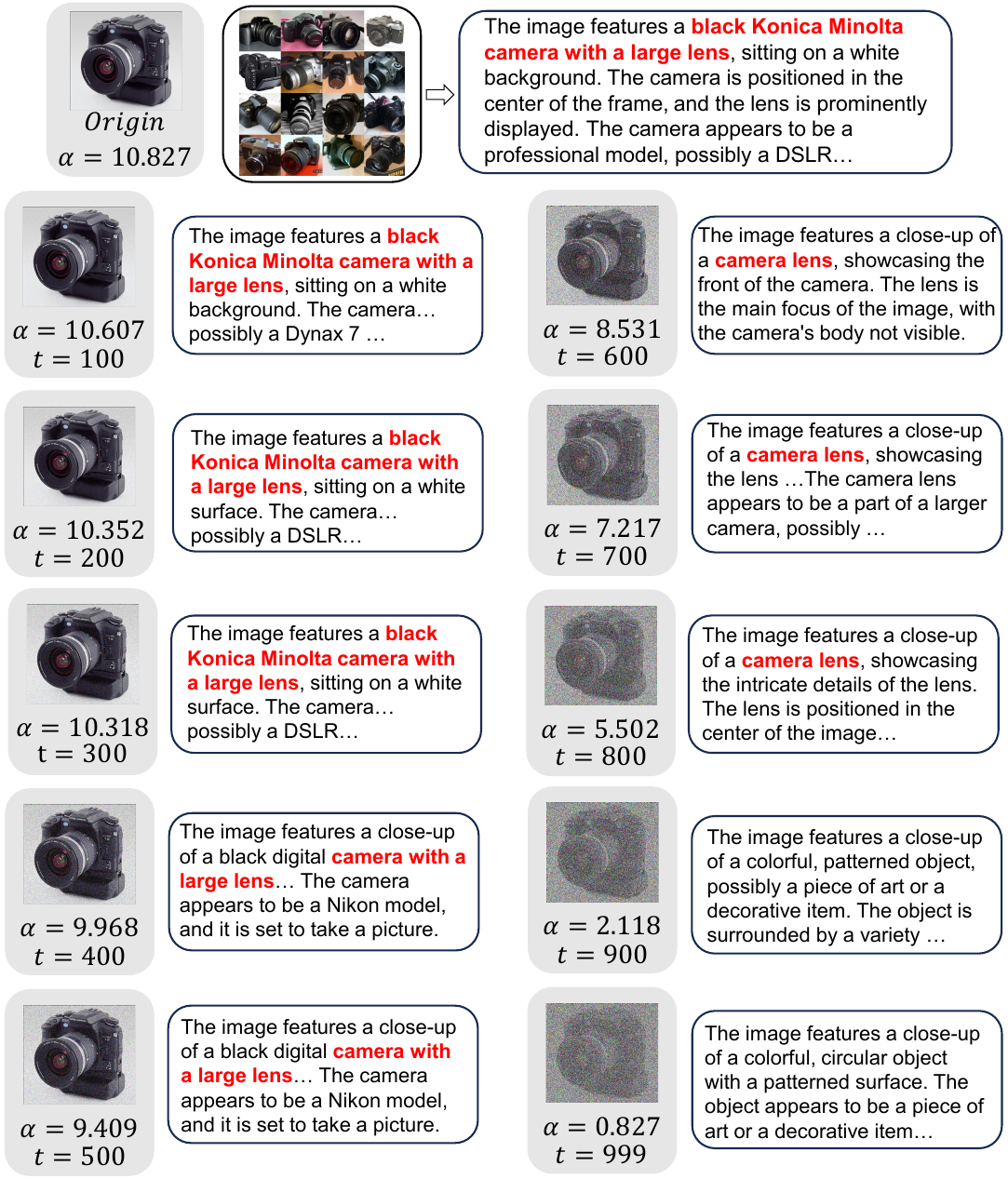}
    \caption{Example of noise affecting visual feature representations. The image contains a camera. As noise increases, the activation of the ``camera'' neuron gradually decreases, and the LVLM output progressively loses detail: from ``black Konica Minolta camera with a large lens'' to ``camera with a large lens,'' and finally no camera description. This demonstrates how noise disrupts internal semantic representations, leading to hallucinations. It also highlights the advantage of SAEs in decoupling dense LVLM features into sparse neurons, allowing us to track and analyze internal visual feature changes at the neuron level.}
    \vspace{-5pt}
    \label{fig:add_noise_vis}
\end{figure*}

To illustrate this phenomenon more intuitively, \Cref{fig:add_noise_vis} shows an example image containing a camera. As increasing levels of noise are applied, the activation of the ``camera'' neuron gradually diminishes. Correspondingly, the LVLM output exhibits a progressive semantic drift: initially describing a ``black Konica Minolta camera with a large lens,'' then simplifying to ``camera with a large lens,'' and eventually omitting the camera entirely. This case demonstrates how noise-induced disruptions at the neuron level directly erode semantic fidelity in visual features, ultimately manifesting as hallucinations in model outputs.

Importantly, this example underscores the value of SAEs: by decomposing dense visual embeddings into sparse neurons, we gain the ability to trace how specific semantic concepts evolve under perturbations. This neuron-level perspective provides interpretability and analytical clarity, enabling us to pinpoint which neurons are destabilized and how this relates to output degradation. Such insights establish a principled foundation for designing targeted interventions to mitigate hallucinations and improve LVLM reliability.

\clearpage

\section{Complex and Diverse Case Studies for Neuron-Level Steering in LVLMs}

We present a range of challenging case studies to demonstrate how neuron-level steering enables fine-grained, interpretable control over LVLM outputs. These examples illustrate not only the feasibility of manipulating specific concepts but also how scene complexity, semantic distribution, and neuron hierarchy affect intervention difficulty and outcomes.

\begin{figure*}[t]
    \centering
    \includegraphics[width=0.95\linewidth]{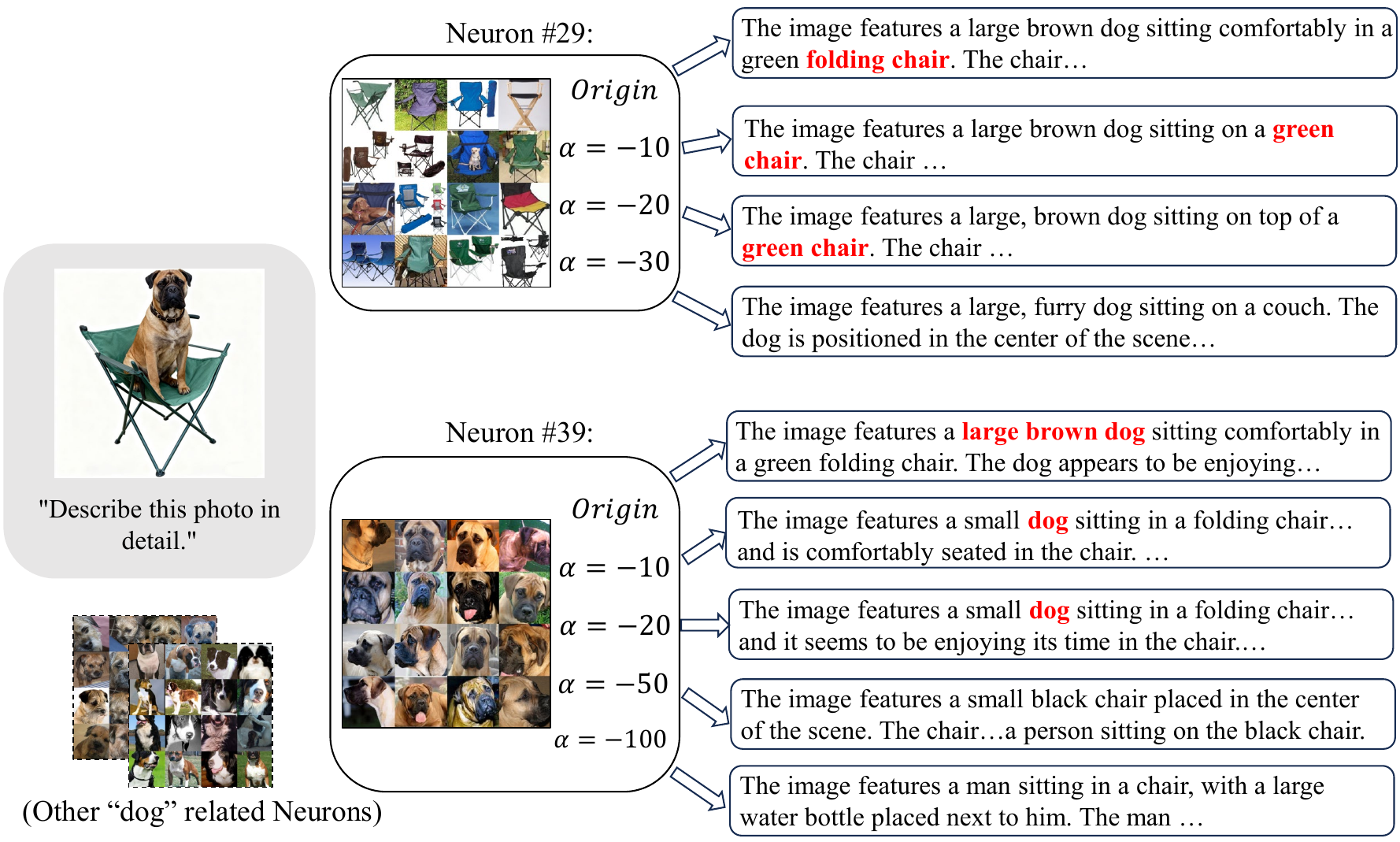}
    \caption{Multi-concept suppression. Suppressing ``chair'' neurons effectively removes chairs from model outputs. Suppressing ``dog'' neurons is more challenging, requiring stronger intervention since the SAE has learned a hierarchy of dog-related concepts (e.g., different breeds). This highlights the difficulty of eliminating concepts encoded in multiple fine-grained neurons.}
    \vspace{-5pt}
    \label{fig:steering_two_neurons}
\end{figure*}

\paragraph{Multi-Concept Suppression.} 
\Cref{fig:steering_two_neurons} depicts a scene containing multiple objects (dog and chair). By selectively suppressing neurons corresponding to each object, we can remove them from generated captions or descriptions in a controlled manner. Interestingly, suppression difficulty varies across concepts. For example, removing ``chair'' neurons is relatively straightforward, with a weight of $\alpha=-30$ sufficient to eliminate chairs from outputs. In contrast, ``dog'' neurons require much stronger intervention, sometimes leaving residual references until $\alpha=-100$ is applied. Analysis reveals that SAE encodes a hierarchical and distributed representation for dog-related concepts, including multiple neurons for different breeds, poses, and contextual cues. This demonstrates that the difficulty of neuron-level control is directly linked to how a concept is represented internally and distributed across neurons.

\begin{figure*}[t]
    \centering
    \includegraphics[width=0.95\linewidth]{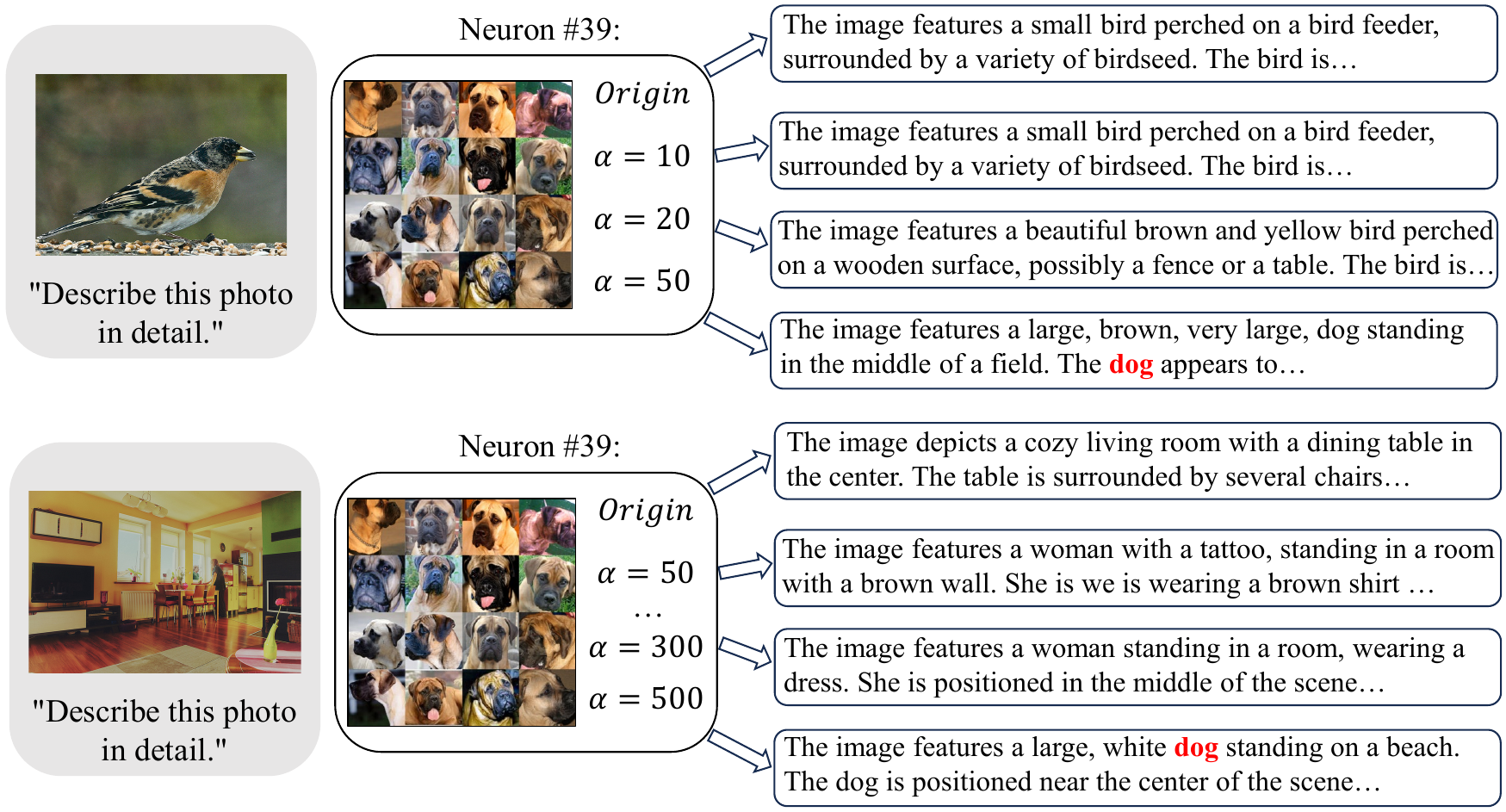}
    \caption{Concept insertion in simple contexts. By slightly amplifying a single dog-related neuron, the model begins to hallucinate the presence of dogs in unrelated scenes. Compared to suppression, concept insertion is easier: small weights suffice to introduce the new concept.}
    \vspace{-5pt}
    \label{fig:steering_dog_neurons}
\end{figure*}

\paragraph{Concept Insertion in Simple and Complex Contexts.} 
We also study the insertion of concepts into scenes of varying complexity. As shown in \Cref{fig:steering_dog_neurons}, inserting a dog concept into a simple bird-dominated scene requires only a modest weight ($\alpha=50$) for the concept to appear. However, in a more complex scene with multiple objects, significantly larger weights ($\alpha=500$) are needed for the concept to manifest reliably. These results highlight that the effectiveness of neuron-level interventions is sensitive to scene complexity and competition among activated neurons. Simpler scenes allow easier manipulation, whereas complex scenes require careful tuning to overcome interference from competing visual features.

\begin{figure*}[t]
    \centering
    \includegraphics[width=0.95\linewidth]{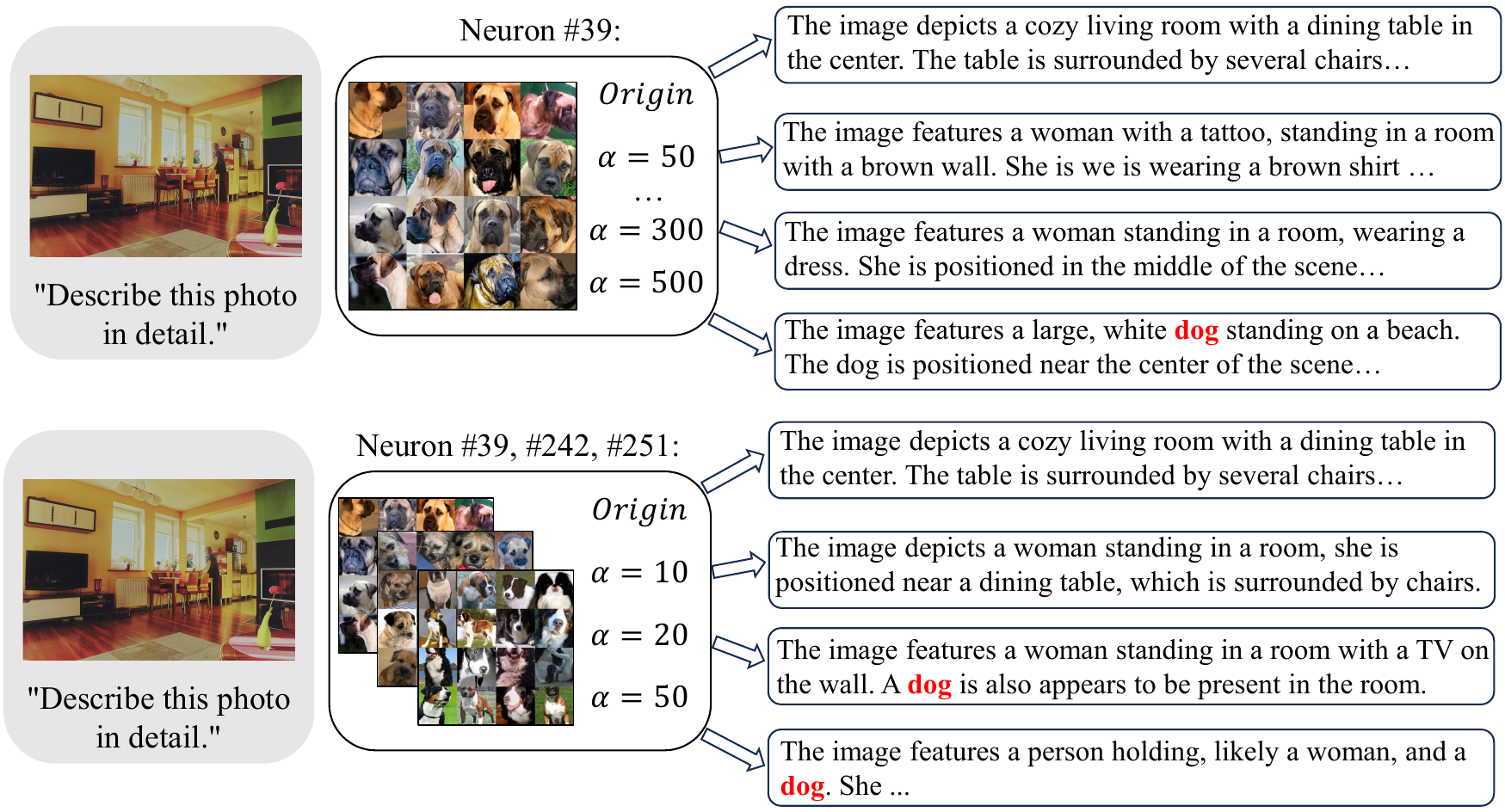}
    \caption{Concept insertion in complex contexts. (a) Steering with a single dog-related neuron requires a very large weight ($\alpha=500$) to produce visible effects. (b) Coordinated steering of three dog-related neurons with smaller weights ($\alpha=20$ each) yields natural insertions. This demonstrates the advantage of multi-neuron steering and motivates our CNS approach.}
    \vspace{-5pt}
    \label{fig:steering_single_multi_insert}
\end{figure*}

\begin{figure*}[t]
    \centering
    \includegraphics[width=0.95\linewidth]{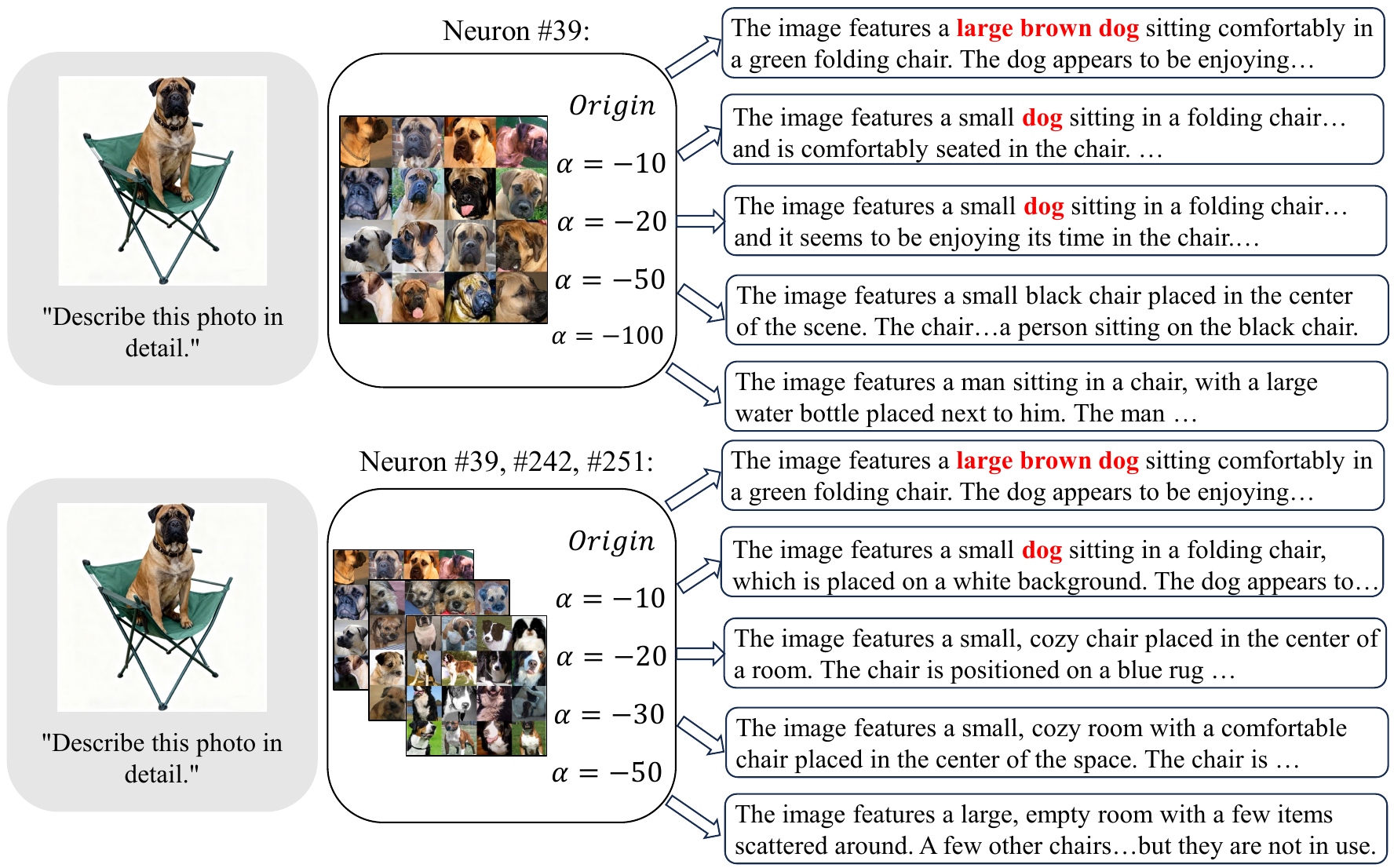}
    \caption{Concept suppression in complex contexts. (a) Suppressing a single dog-related neuron requires a very large negative weight ($\alpha=-100$) before the concept disappears from outputs. (b) Jointly suppressing three dog-related neurons with smaller weights ($\alpha=-10$ each) removes the concept more naturally and reliably, illustrating the effectiveness of multi-neuron steering.}
    \vspace{-5pt}
    \label{fig:steering_single_multi_supress}
\end{figure*}

\paragraph{Multi-Neuron Steering in Complex Scenes.} 
In highly complex contexts, steering a single neuron often requires extreme weights before a concept appears in outputs (\Cref{fig:steering_single_multi_insert}). Coordinated adjustment of multiple concept-related neurons with smaller individual weights produces more stable, natural, and reliable insertions. A similar pattern holds for suppression (\Cref{fig:steering_single_multi_supress}): targeting a single neuron requires very large negative weights, while modulating multiple neurons simultaneously with smaller magnitudes removes concepts more effectively. These observations underline the advantages of multi-neuron steering for both insertion and suppression, providing robustness against distributed representations and ensuring smoother, more predictable outputs. Such findings directly motivate our CNS framework, which automatically identifies and adjusts multiple neurons for reliable fine-grained control.

\paragraph{Insights into Internal Representations.} 
Across all cases, neuron-level steering provides interpretable insights into how visual concepts are internally represented in LVLMs. Concepts are often encoded across several neurons with varying sensitivities, and scene complexity, occlusion, or overlapping features can lead to partial activation of irrelevant neurons, contributing to hallucinations. Multi-neuron interventions reveal the distributed and hierarchical nature of these representations and offer a principled way to correct or manipulate outputs. Importantly, these analyses highlight how CNS leverages contrastive identification of image-specific neurons to selectively enhance or suppress concept-relevant activations, ensuring controllable behavior while preserving overall scene understanding.

\paragraph{Summary.} 
Overall, these studies demonstrate that neuron-level steering offers a powerful, interpretable, and generalizable approach for controlling LVLM behavior. Its effectiveness depends on factors such as semantic distribution across neurons, scene complexity, and concept hierarchy. By revealing the internal mechanisms underlying hallucinations and providing actionable intervention strategies, these case studies highlight how CNS can mitigate hallucinations and, more fundamentally, enhance the reliability of visual feature representations. This work provides practical guidance for fine-grained, concept-level, and interpretable control of LVLM outputs.

\clearpage

\section{Discussion and Comparison with Register Neurons}

Several phenomena in LVLMs exhibit stable, high-norm, input-invariant activations, including \emph{always-on neurons}, \emph{register neurons}~\cite{register_need,register_dont_need}, \emph{massive activations}~\cite{sun2024massive_activations_attention_sinks}, and \emph{attention sinks}~\cite{xiao2023efficient_attention_sinks_ori,kang2025see_attention_sinks}. These phenomena share certain characteristics, such as persistently high activation magnitudes and minimal dependence on specific local visual content. In our observations, always-on neurons typically exhibit sparse, high-magnitude activations between 10--80, while most of the top-40 neurons fall in the 5--15 range. They appear consistently across inputs and primarily correspond to non-core, global features, reflecting decoupled latent factors in the internal representation.

\textbf{Shared characteristics: sparse, input-invariant, globally stable activations}
\begin{itemize}
    \item \textbf{Activation sparsity:} selectively active in specific neurons, facilitating sparse and decoupled representations.
    \item \textbf{Input-independence:} activation patterns are largely independent of specific local visual content.
    \item \textbf{Global role:} capture high-level, non-local computations or statistical factors within the model.
\end{itemize}

Despite these similarities, always-on neurons differ from register neurons in several aspects:
\begin{itemize}
    \item \textbf{Generation mechanism:} Register neurons arise from MLP outputs within a layer, whereas always-on neurons are identified via SAE decomposition from the entire layer output, capturing sparse latent factors across the full representation.
    \item \textbf{Distribution and consistency:} Register neurons vary in number and activation location across inputs. In contrast, always-on neurons consistently appear in the same set of neurons across nearly all inputs, hence ``always-on''.
    \item \textbf{Analysis and interpretability:} Always-on neurons provide directly visualizable, decoupled representations, whereas register neurons are primarily interpreted indirectly through their influence on model outputs.
    \item \textbf{Concept-level controllability:} The sparsity and decoupling of always-on neurons enable fine-grained interventions that can modulate specific concept representations without affecting unrelated features. This property is unique to always-on neurons and is not observed in register neurons or other high-norm activations.
\end{itemize}

Overall, always-on neurons encode sparse, decoupled latent factors that expose the internal feature structure in a directly interpretable manner. Their sparsity and decoupling further enable fine-grained, concept-level interventions, allowing specific internal representations to be modulated without affecting unrelated features. In contrast, register neurons are structural components tied to MLP parameters and do not provide the same level of sparsity, interpretability, or controllability.

\textbf{Future directions:} Investigating the relationships among always-on neurons, register neurons, massive activations, and attention sinks could yield deeper insights into persistent high-norm activations in LVLMs. In particular, analyzing SAE factors from a sparsity- and norm-based perspective may elucidate the structural and functional roles of these neurons, their contribution to decoupled and interpretable internal representations, and their potential for concept-level interventions in large vision-language models.

\end{document}